%% file: acl_latex.tex
\pdfoutput=1

\documentclass[11pt]{article}

\usepackage[preprint]{acl}

\usepackage{times}
\usepackage{latexsym}

\usepackage[T1]{fontenc}

\usepackage[utf8]{inputenc}

\usepackage{microtype}

\usepackage{inconsolata}

\usepackage{graphicx}

\usepackage{tabularx}
\usepackage{multirow}
\usepackage{multicol}
\usepackage{url}
\usepackage{booktabs}
\usepackage{xcolor}
\usepackage{colortbl}
\usepackage{algorithm}
\usepackage{algorithmic}
\usepackage{tcolorbox}
\usepackage{pifont}
\usepackage{amsmath}
\usepackage{amsthm}
\usepackage{amssymb}
\usepackage{relsize}

\usepackage{enumitem}
\usepackage{enumerate}
\usepackage{xspace}
\usepackage{subcaption}
\definecolor{lightgreen}{rgb}{0.88, 1, 0.88}
\definecolor{lightpurple}{RGB}{180, 150, 255}
\definecolor{lightred}{RGB}{255, 204, 203}
\definecolor{lightblue}{RGB}{173, 216, 230}
\usepackage{xcolor}

\newcommand{\suhang}[1]{\textcolor{blue}{#1}}

\newcommand{\minhua}[1]{\textcolor{purple}{#1}}
\newcommand{\method}{\ensuremath{\textnormal{\textsc{TESSA}}}\xspace}
\newcommand{\nop}[1]{}

\title{Decoding Time Series with LLMs: A Multi-Agent Framework\\ for Cross-Domain Annotation}

\author{Minhua Lin$^{1}$\thanks{Work done during an internship at NEC Labs America.}, Zhengzhang Chen$^{2}$\thanks{Corresponding author.}, Yanchi Liu$^2$, Xujiang Zhao$^2$,\\ \textbf{Zongyu Wu}$^{1}$, \textbf{Junxiang Wang}$^2$, \textbf{Xiang Zhang}$^1$, \textbf{Suhang Wang}$^1$, \textbf{Haifeng Chen}$^{2}$   \\  
 $^{1}$The Pennsylvania State University $^{2}$NEC Laboratories America\\
\texttt{\{mfl5681,zongyuwu,xzz89,szw494\}@psu.edu}\\ 
\texttt{\{zchen,yanchi,xuzhao,junwang,haifeng\}@nec-labs.com} \\
 }

\begin{document}
\maketitle

\input{0_abstract}

\input{1_intro}
\input{5_related_work}
\input{3_methodology}

\input{4_experiment}
\input{6_conclusion}

\bibliography{acl_latex}

\clearpage
\appendix
\input{7_appendix}

\end{document}

%% file: 0_abstract.tex
\begin{abstract}

Time series data is ubiquitous across various domains, including manufacturing, finance, and healthcare. High-quality annotations are essential for effectively understanding time series and facilitating downstream tasks. However, obtaining such annotations is challenging, particularly in mission-critical domains. In this paper, we propose \method, a multi-agent system designed to automatically generate both general and domain-specific annotations for time series data. \method introduces two agents: a general annotation agent and a domain-specific annotation agent. The general agent captures common patterns and knowledge across multiple source domains, leveraging both time-series-wise and text-wise features to generate general annotations. Meanwhile, the domain-specific agent utilizes limited annotations from the target domain to learn domain-specific terminology and generate targeted annotations. Extensive experiments on multiple synthetic and real-world datasets demonstrate that \method effectively generates high-quality annotations, outperforming existing methods. 

\end{abstract}

%% file: 1_intro.tex
\section{Introduction}






Time series data is prevalent 
in various fields such as manufacturing~\cite{hsu2021multiple}, finance~\cite{lee2024survey}, and healthcare~\cite{cascella2023evaluating}.  
It captures critical temporal patterns essential for informed decision-making. However, general users frequently encounter difficulties in interpreting this data due to its inherent complexity, particularly in multivariate contexts where multiple variables interact over time. 
Furthermore, effective interpretation typically requires domain-specific knowledge to properly contextualize these patterns, thereby posing significant challenges for individuals without specialized expertise.


High-quality annotations are crucial for addressing these interpretive challenges. Annotations provide meaningful context or insights into time series data, highlighting important patterns, events, or anomalies. They facilitate accurate analysis, forecasting, and decision-making, enhancing the performance of downstream tasks such as anomaly detection, trend prediction, and automated reporting. For instance, in predictive maintenance, understanding sensor data trends is vital for preventing equipment failure, while in finance, interpreting stock price movements is crucial for informed investment strategies. Despite their importance, high-quality annotations are often scarce in real-world applications. This scarcity stems primarily from the reliance on domain experts for manual annotation, which is resource-intensive, costly, and prone to inconsistencies. Moreover, the need for precise and domain-specific terminology further complicates the annotation process, as different fields require highly specialized knowledge for accurate and contextually relevant interpretation.
\begin{figure}[t]
    \centering
    \includegraphics[width=0.85\linewidth]{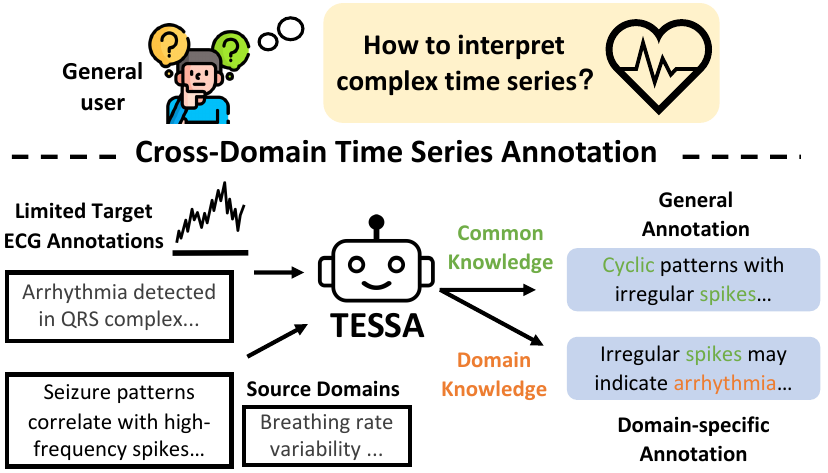}
    \caption{How to annotate time series across different domains?}
    \label{fig:cross_domain_time_series_annotation}
\end{figure}

To alleviate the above issues, one straightforward approach is to leverage external resources to generate annotations~\cite{liu2024time}. For example, Time-MMD~\cite{liu2024time} uses web searches to retrieve information as annotations, aiming to find similar patterns and descriptions from the internet. Others~\cite{jin2024time,liu2024unitime} directly adopt large language models (LLMs) for annotation, leveraging LLMs' great language understanding capability. Prototype-based methods, such as prototype networks~\cite{ni2021interpreting}, have also been employed to identify representative examples for annotation. However, \textit{these methods often fall short of producing high-quality annotations}. Web search-based methods may retrieve irrelevant or inconsistent information. LLMs, while powerful, tend to generate generic annotations, capture only basic patterns, or even hallucinate, and fail to account for the complex nature of time series data. Prototype networks rely on large amounts of data to train the network and identify representative prototypes, but the scarcity of high-quality annotations limits the quality and representativeness of these prototypes, making it difficult to generalize effectively to new or unseen patterns.

To address these limitations, we propose to extract knowledge from existing annotations across multiple source domains and transfer this knowledge to target domains with limited annotations. Specifically, as shown in Fig.~\ref{fig:cross_domain_time_series_annotation}, we aim to develop a system that automatically interprets time series data across various fields using common or domain-specific language. Formally, given abundant annotations from multiple source domains and limited annotations from a target domain, our goal is to leverage both time-series-wise and text-wise knowledge to generate accurate and contextually appropriate annotations for the target domain. There are two major technical challenges in developing such a system: \textbf{(i)} How to extract common knowledge from source domains? \textbf{(ii)} How to learn domain-specific jargon from limited target-domain annotations? 

To tackle these challenges and overcome the limitations of existing methods, we propose \textbf{TESSA}, a multi-agent system designed for both general and domain-specific \underline{T}im\underline{E} \underline{S}erie\underline{S} \underline{A}nnotation. As illustrated in Fig.~\ref{fig:TESSA_overall_framework}, \method introduces two agents: a general annotation agent and a domain-specific annotation agent. The general annotation agent focuses on capturing common patterns and knowledge across various domains to generate annotations understandable by general users. To learn common knowledge from multiple domains, the general agent employs a time series-wise feature extractor and a text-wise feature extractor to extract both time-series-wise and text-wise features from time series data and domain-specific annotations from multiple source domains. To ensure important features are included in the general annotations, two feature selection methods—LLM-based and reinforcement learning-based selection—are introduced to effectively and efficiently select both the top-$k$ most important time-series-wise and text-wise features. The domain-specific agent leverages limited target-domain annotations to learn and generate annotations for specific domains using domain-specific terminologies (jargon). It incorporates a domain-specific term extractor to learn jargon from the limited target-domain annotations. Additionally, an annotation reviewer is proposed to maintain consistency between general annotations and domain-specific annotations.

Our contributions are: (i) \textbf{Problem}. 
    We explore a novel problem in cross-domain multi-modal time series annotation, bridging the gap between general understanding and domain-specific interpretation; (ii) \textbf{Framework}. We propose a novel multi-agent system, \method, designed for both general and domain-specific time series annotation by leveraging both time-series-wise and text-wise knowledge from multiple domains; 
    (iii) \textbf{Experiments}. Extensive experiments on multiple synthetic and real-world datasets demonstrate the quality of the general and domain-specific annotations generated by \method.

%% file: 5_related_work.tex
\section{Related Work}

\begin{figure*}[t]
    \centering
    \includegraphics[width=0.9\linewidth]{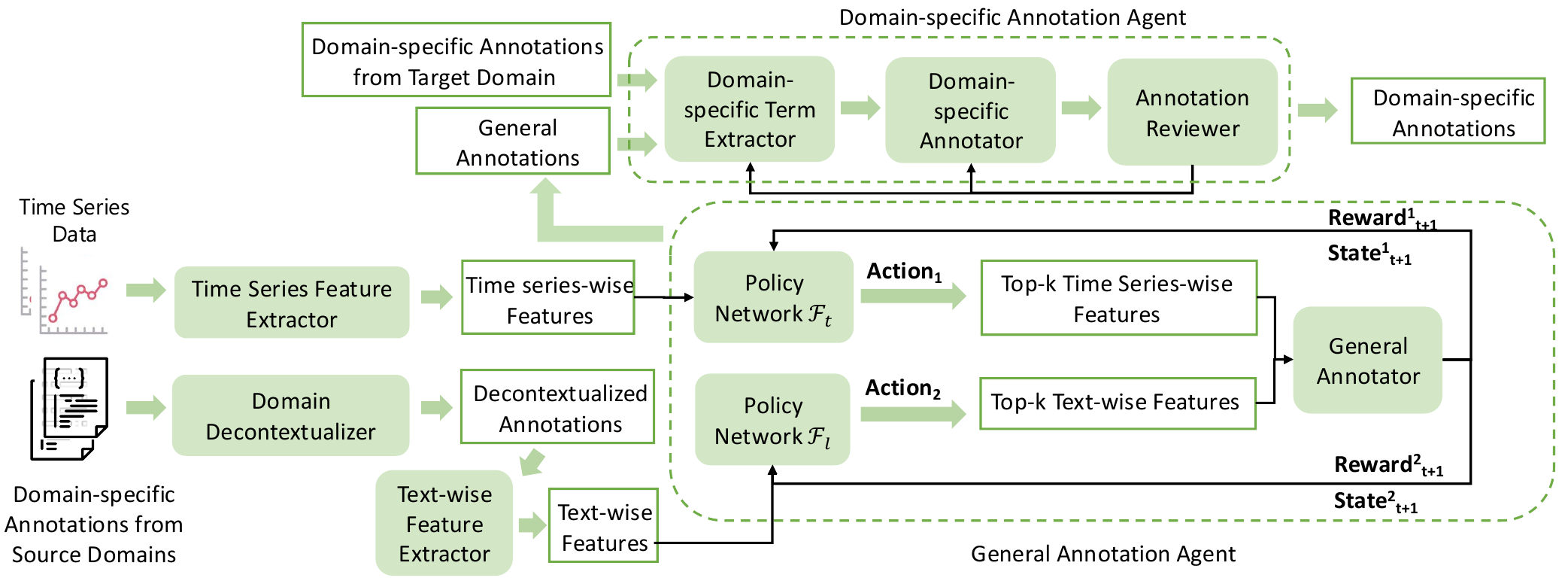}
   \vskip -0.5em
    \caption{Overall framework of \method. It consists of two main agents: a general annotation agent, which generates domain-independent annotations by selecting salient time-series and textual features, and a domain-specific annotation agent, which refines these annotations by incorporating domain-specific terminology.}
    \label{fig:TESSA_overall_framework}
\end{figure*}


\textbf{Time Series Annotation}.
Time series annotation aims to assign labels or descriptions to specific segments, events, or patterns within a time series dataset to highlight significant features for further analysis. Traditionally, this process has relied on manual annotation~\cite{reining2020annotation}, which is often time-consuming, labor-intensive, and requires substantial domain expertise. To reduce the effort needed for creating large-scale, high-quality annotated datasets, several studies have proposed semi-automatic annotation approaches~\cite{cruz2019sensor, Nino2016ScalableSA} that require minimal manual input or post-annotation revisions. Despite these advancements, fully automated time series annotation remains underexplored due to the challenges of capturing semantic and contextual information from the data~\cite{Yordanova2018creating}.

\noindent\textbf{LLMs for Time Series Analysis}.
Recent advancements in LLMs have showcased their strong capabilities in sequential modeling and pattern recognition, opening up promising new directions for time series analysis. Several studies~\cite{xue2023promptcast, yu2023temporal, gruver2024large, jin2024time, li2024frozen} have explored how LLMs can be effectively leveraged in this context. For instance, PromptCast~\cite{xue2023promptcast} is a pioneering work that applies LLMs to general time series forecasting using a sentence-to-sentence approach. Yu \textit{et al.} \cite{yu2023temporal} extend this by investigating the application of LLMs to domain-specific tasks, such as financial time series forecasting. LLMTime~\cite{gruver2024large} demonstrates the efficacy of LLMs as time series learners by employing text-wise tokenization to represent time series data. Time-LLM~\cite{jin2024time} reprograms time series data into textual prototypes for input into LLaMA-7B, enriched with natural language prompts that include domain expert knowledge and task-specific instructions. More details of related works are in Appendix~\ref{appendix:related_works_llm_TSA}.

\noindent\textbf{Cross-modality Knowledge Transfer Learning through Pre-trained Models}. 
There has been growing interest in leveraging pre-trained models for cross-modality knowledge transfer, particularly between the language, vision, and time series domains~\cite{bao2022vlmo,lu2022frozen,yang2021voice2series,zhou2023one}. For instance, \citet{bao2022vlmo} proposes a stagewise pre-training strategy that trains a language expert using frozen attention blocks pre-trained on image-only data. Similarly, \citet{lu2022frozen} examines the transferability of language models to other domains, while \citet{zhou2023one} applies pre-trained language and image models to time series analysis tasks. To the best of our knowledge, no previous work has specifically explored cross-modality knowledge transfer for time series annotation. Our work aims to fill this gap by investigating the application of cross-modality transfer learning in the context of automatic time series annotation.


%% file: 3_methodology.tex
\section{Methodology}
\label{sec:methdology}

In this section, we define the problem and present the details of our proposed \method framework, which aims to generate both general and domain-specific annotations for time series data. 

\noindent\textbf{Cross-Domain Time Series Annotation Problem}. Given several source domains $\{\mathcal{D}_{s_1}, \mathcal{D}_{s_2}, \ldots\}$ and a target domain $\mathcal{D}_t$, let $\{e^1_{s_i}, e^2_{s_i}, \ldots\}$ denote the domain-specific annotations 
from the source domain $\mathcal{D}_{s_1}$, and $\{e^1_{t}, e^2_{t}, \ldots\}$ represent the limited domain-specific annotations from the target domain $\mathcal{D}_t$. Suppose $\mathbf{X} = (\mathbf{x}_1, \cdots, \mathbf{x}_L)$ is a time series in $\mathcal{D}_t$, where $L$ is the number of past timestamps and $\mathbf{x}_i = (x_{1i}, \cdots, x_{Ci})^{T} \in \mathbb{R}^C$ represents the data from $C$ different channels at timestamp $i$. The objective of cross-domain time series annotation is to generate the general annotation $e_g$ and the domain-specific annotation $e_s$ for $\mathbf{X}$ based on the annotations from both the source and target domains. More notations are provided in Appendix~\ref{appendix:notations}.

\noindent\textbf{Overview of TESSA}. As illustrated in Fig.~\ref{fig:TESSA_overall_framework}, the proposed \method comprises two key components: a general annotation agent and a domain-specific annotation agent. The general annotation agent is responsible for generating domain-independent annotations and consists of several modules: a time series feature extraction module to capture time-series-specific features, a domain decontextualization module to 
convert domain-specific text into common language, 
a text feature extraction module to retrieve textual features from the decontextualized text, two policy networks for selecting the top-$k$ most salient time-series and textual features, and a general annotator to produce general annotations based on the selected features. The domain-specific annotation agent refines the general annotations to generate domain-specific annotations. It includes a domain-specific term extractor to identify key terminology from a limited set of target-domain annotations and a domain-specific annotator to adjust the general annotations accordingly. An annotation reviewer further enhances the quality of the domain-specific annotations. Next, we introduce details of each component.

\subsection{Multi-modal Feature Extraction}
To address the challenge of extracting common knowledge from source domains, we introduce two feature extraction modules: a time-series feature extractor and a text-wise feature extractor, which extract features from time series data and source-domain annotations. We also propose a domain decontextualizer to enhance the extraction of common knowledge from multi-source annotations.

\nop{\noindent \textbf{text-series Feature Extraction}. We develop a time series extraction toolbox $f_t$ to extract various text-series features from given time series data. Formally, for each channel $c\in C$, the set of text-series features $\mathbf{F}_t$ is denoted as
\begin{equation}
    \mathbf{F}_t = \{f_t^1,\cdots,f_t^{n_t}\} = \mathcal{M}_{r}(\mathbf{X}),
\end{equation}
where $f^i_{t}$ is the $i$-th extracted text-series features of $\mathbf{X}$, $n_t$ is the number of extracted features. Additionally, for multivariate time series data, some inter-variable features (e.g., Pearson correlation) are also involved. More details of the feature extraction are in Appendix~\ref{appendix:more_details_time_series_feature_extraction}.}

\noindent \textbf{Time Series Feature Extraction}. 
We extract features from time series data through a structured process $\mathcal{M}_{r}$. Formally, for each channel $c \in C$, the set of time-series features $\mathbf{F}_t$ is denoted as:
{\begin{equation}\small
    \mathbf{F}_t = \{f_t^1,\cdots,f_t^{n_t}\} = \mathcal{M}_{r}(\mathbf{X}),
\end{equation}}
where $\mathcal{M}_r$ denotes the feature extraction framework applied to $\mathbf{X}$, $f^i_{t}$ is the $i$-th extracted feature of $\mathbf{X}$, and $n_t$ is the number of extracted features. For multivariate time series data, inter-variable features (\textit{e.g.}, Pearson correlation) are also included. Details of $\mathcal{M}_r$ are provided in Appendix~\ref{appendix:more_details_time_series_feature_extraction}.

\nop{Besides the text-series features, the textual annotations from source domains usually contain rich knowledge that benefit the interpreting time series data in the target domain.

To extract these text common knowledge from source domains, one naive method is to directly extract text-wise features by using LLMs from domain-specific annotations thanks to the rich real-world knowledge of LLMs. However, many real-world domains lack sufficient high-quality annotations to extract insightful time series features, and the presence of too many domain-specific terminologies will further hinder the extraction of these features.}

\noindent \textbf{Domain Decontextualization}. In addition to time-series-wise features, textual annotations from source domains often contain valuable information (such as support or resilience in finance time series annotations) for interpreting time series data. A straightforward method to extract this knowledge is to use LLMs on domain-specific annotations, leveraging their real-world knowledge. However, in practice, many domains lack sufficient high-quality annotations, and domain-specific terminology can further hinder effective extraction.

\nop{To solve these issues and facilitate the transfer of knowledge from source domains to target domains, 
a domain decontextualization LLM is introduced to convert domain-specific annotations from multiple source domains into general annotations by removing domain-specific terminology, making it easier to extract common knowledge across various domains. Specifically, given an domain-specific annotation $e_s$ in the domain $d_i$, a domain decontextualization LLM will give the decontextualized annotation $e_d$ based on $e_s$ and $d_i$ as
\begin{equation}
    e_d=\mathcal{M}_d(p_{de}(e_s,d_i)),
\end{equation}
where $\mathcal{M}_d$ is the domain decontextualization LLM. More details including the prompt template of $p_{de}$ and an example can be found in Appendix~\ref{appendix:more_details_domain_decontextualization}.}

To address these challenges and facilitate knowledge transfer from source to target domains, we introduce a domain decontextualization LLM to convert domain-specific annotations into general annotations by removing domain-specific terminology. This makes it easier to extract common knowledge across domains. Specifically, given a domain-specific annotation $e_s^i$ in domain $d_i$, the decontextualized annotation $e_d^i$ is obtained as:
\begin{equation}
\small
    e_d^i = \mathcal{M}_d(p_{de}(e_s^i, d_i)),
\end{equation}
where $\mathcal{M}_d$ is the domain decontextualization LLM. 


\nop{\noindent \textbf{text-wise Feature Extraction}. After domain decontextualization, we introduce an LLM $\mathcal{M}_{l}$ as a text-wise feature extractor to extract text common knowledge from multiple source domains that are not explored from the text-series. Formally, given a set decontextualized annotations $\{e_d^i\}_{i=1}^{n_{d}}$ and a text-wise feature extractor $\mathcal{M}_{l}$, the extracted text-wise features are denoted as
\begin{equation}
    \mathbf{F}_l = \{f_l^1,\cdots,f_l^n\} = \mathcal{M}_{l}(p_{l}(\{e_d^i\}_{i=1}^{n_{d}})),
\end{equation}
where $p_{l}$ is the prompt of text-wise feature extraction. The prompt template of $p_{l}$ and an example of extracting text-wise feature can be found in Appendix~\ref{appendix:more_detail_text_wise_extraction}.}

\noindent \textbf{Text Feature Extraction}. After decontextualization, we use an LLM $\mathcal{M}_{l}$ to extract textual features from multiple source domains. Formally, given a set of decontextualized annotations $\{e_d^i\}_{i=1}^{n_{d}}$ and the text feature extractor $\mathcal{M}_{l}$, the extracted textual features are denoted as:
\begin{equation}
\small
    \mathbf{F}_l = \{f_l^1,\cdots,f_l^n\} = \mathcal{M}_{l}(p_{l}(\{e_d^i\}_{i=1}^{n_{d}})),
\end{equation}
where $p_{l}$ is the prompt for text feature extraction to guide $\mathcal{M}_{l}$ to output the text-wise features explicitly or implicitly mentioned in the decontextualized annotations. 


\subsection{Adaptive Feature Selection}
\label{sec:adaptive_feature_selection}
With a diverse set of features extracted from time series and text data, it becomes essential to focus on the most relevant ones to ensure the generated annotations remain concise and interpretable. Moreover, repeatedly querying LLMs with both the old and new data\footnote{To avoid redundancy, unless specified otherwise, `data' in this paper refers to time series and their corresponding textual annotations from various domains.} each time wastes computational resources and incurs additional costs, especially when using non-open-source models.


To address these issues, we propose a hybrid strategy for adaptive feature selection that combines \textit{Offline LLM-based Feature Selection} with \textit{Incremental Reinforcement Learning-based Feature Selection}. The incremental method builds on the offline approach, minimizing the need to re-query LLMs with both old and new data as it arrives.




\noindent \textbf{Offline LLM-based Feature Selection}. Leveraging LLMs' reasoning abilities, we introduce a feature selection method using LLM-generated feature importance scores to identify the top-$k$ most important time-series-wise and text-wise features. Features mentioned more frequently—either explicitly or implicitly—in annotations are assigned higher importance scores.

Specifically, given an LLM as the feature selector $\mathcal{M}_{sel}$, we prompt $\mathcal{M}_{sel}$ with domain-decontextualized annotations $\{e_d^i\}_{i=1}^{n_d}$ and the extracted features $\{f_t^i\}_{i=1}^{n_t}$ and $\{f_l^i\}_{i=1}^{n_l}$ to generate numerical feature importance scores: $\mathbf{s}_t = [s_1,\cdots,s_{n_t}]$ for time-series-wise features and $\mathbf{s}_l = [s_1,\cdots,s_{n_l}]$ for text-wise features.
\begin{equation}
\small
\label{eq:score_llm_select}
\begin{aligned}
    s_j = \mathcal{M}_{sel}({p}_{{score}}(f_t^j, \{e_d^i\}_{i=1}^{n_d})), \ \ \forall{j}\in\{1,\cdots,n_t\}, \\
    s_k= \mathcal{M}_{sel}({p}_{{score}}(f_l^k, \{e_d^i\}_{i=1}^{n_d})), \ \ \forall{k}\in\{1,\cdots,n_l\}, \\
\end{aligned}
\end{equation}
Here, $p_{score}$ is the prompt used to score feature importance. Higher scores, $s_j$ and $s_k \in \mathbb{R}^{+}$, indicate that the features $f_t^j$ and $f_l^k$ appear more frequently, either explicitly or implicitly, in the domain-decontextualized annotations $\{e_d^i\}_{i=1}^{n_d}$. 
To ensure that explicitly mentioned features receive higher importance scores, we instruct $\mathcal{M}_{sel}$ to assign greater weight to features that are explicitly referenced in the annotations. More details are provided in Appendix~\ref{appendix:more_details_llm_based_feature_selection}.

\nop{Specifically, given an LLM as a feature selector ($\mathcal{M}_{sel}$), we prompt it with domain-decontextualized annotations and extracted features to obtain feature importance scores:
\begin{equation}
\small
\begin{aligned}
    s_j = \mathcal{M}_{sel}({p}_{{score}}(f_t^j, \{e_d^i\}_{i=1}^{n_d})), \ \ \forall{j}\in\{1,\cdots,n_t\}, \\
    s_k= \mathcal{M}_{sel}({p}_{{score}}(f_l^k, \{e_d^i\}_{i=1}^{n_d})), \ \ \forall{k}\in\{1,\cdots,n_l\}, \\
\end{aligned}
\end{equation}
Higher scores indicate more frequent or explicit mentions in the annotations.}


\noindent \textbf{Incremental Reinforcement Learning-based Feature Selection}. 
When new data\footnotemark[1] arrives, the offline LLM-based approach requires re-querying both old and new data, which becomes burdensome due to LLMs' limited context window. As annotations increase, re-querying all data becomes impractical and costly, leading to higher resource consumption and reduced cost-effectiveness.

To address the limitations of the offline approach, we propose an \textit{Incremental Reinforcement Learning-based Feature Selection} method that is more cost-effective for dynamic environments with evolving data. Specifically, we introduce a multi-agent reinforcement learning (MARL) framework to train two policy networks, $\mathcal{F}_t$ and $\mathcal{F}_l$, to select the top-$k$ most important time-series-wise and text-wise features, respectively. 
These policy networks store knowledge from existing annotations and are incrementally updated as new data arrives. This reduces the need to re-query the LLM with all the data, requiring only the new data during updates.  
As shown in Fig.~\ref{fig:TESSA_overall_framework}, each policy network is initialized with the first three layers of a small LLM, such as GPT-2~\cite{radford2019language}, which remain frozen during training. A trainable multi-head attention layer and a language model (LM) head from GPT-2 follow these layers, using the smallest version of GPT-2 with 124M parameters.

During training, only the multi-head attention layer is updated. For time-series-wise features, given the candidate features $\{f_t^i\}_{i=1}^{n_t}$ and their corresponding feature name tokens $\mathbf{Y}=\{y^i_1,\cdots,y^i_{n_{t}}\}$, the policy network $\mathcal{F}_t$ computes action-values (Q-values) $\mathbf{q}_z = [q_{z, f_t^1}, \cdots, q_{z, f_t^{n_t}}]$ based on the mean logits of the feature names:
\begin{equation} 
\small
\mathbf{q}_s = \mathcal{F}_t(\{y_i\}_{i=1}^{n_t}), 
\end{equation}
A softmax function generates a probability distribution over the features, and the top-$k$ features are selected based on the highest probabilities.

At each timestep, the selected top-$k$ features are passed to the LLM $\mathcal{M}_{sel}$ to obtain their importance scores $s_i, \forall i \in \{1,\cdots,k\}$. The agent receives a reward $r_t$ defined as:

\begin{equation}
\small
\begin{aligned}
    r_t = 
    \begin{cases}
     \sum_{i=1}^{k} s_i, & s_i \geq \tau  \\
    -0.5, & \text{otherwise},
    \end{cases}
\end{aligned}
\end{equation}
where $\tau$ is a threshold to discourage selecting unimportant features. The text-wise feature policy network $\mathcal{F}_l$ undergoes a similar training process.

After training, the policy networks are incrementally updated with only new data, eliminating the need to re-query the LLM with both old and new data. This approach improves the scalability and efficiency of feature selection while reducing computational costs, effectively overcoming the offline approach's limitations. By incrementally updating the policy networks, we ensure that feature selection remains scalable and cost-effective in dynamic environments with evolving data.
More discussion of the necessity of the RL component is provided in Appendix~\ref{sec:discussion_rl_selection}.

\subsection{General Annotation Generation}
After selecting the top-$k$ most important features from both time-series and text, a general annotator is introduced to generate general annotations by analyzing these selected features. An LLM, serving as the general annotator, interprets the given time series data based on the selected features. Formally, given time series data $\mathbf{X} = \{\mathbf{x}_i\}_{i=1}^{L}$ and the selected time-series-wise and text-wise features $\{f_t^i\}_{i=1}^{k_t}$ and $\{f_l^i\}_{i=1}^{k_l}$, the generation of a general annotation $e_g$ is represented as:

\begin{equation}
\small
    e_g = \mathcal{M}_{gen}(p_{gen}(\{\mathbf{x}_i\}_{i=1}^{L}, \{f_t^i\}_{i=1}^{k_t}, \{f_l^i\}_{i=1}^{k_l})),
\end{equation}
where $p_{gen}$ is the prompt for generating general annotations. By emphasizing the signal from the selected common knowledge, the general annotations capture richer patterns that may be overlooked when directly applying LLMs. 

\subsection{Domain-specific Annotation Generation}
Generating domain-specific annotations for time series is crucial as different domains rely on specialized jargon and context-specific terminology to accurately interpret and understand data. Time series data from financial markets, healthcare systems, or industrial processes can exhibit patterns, trends, and anomalies that are unique to each domain. General annotations may overlook critical nuances, whereas domain-specific annotations capture contextual relevance, improving the precision and reliability of downstream analysis or model predictions. By tailoring annotations to a domain's specific lexicon, we can detect meaningful patterns more accurately and make informed decisions.

\noindent \textbf{Domain-specific Term Extractor}. To address the challenge of learning domain-specific terminology, we introduce a domain-specific term extractor. Given limited domain-specific annotations $\{e_t^i\}_{i=1}^{n_{e_t}}$ from the target domain, an LLM $\mathcal{M}_{ext}$ is employed to extract domain-specific terms. We prompt $\mathcal{M}_{ext}$ with the annotations $\{e_t^i\}_{i=1}^{n_{e_t}}$ to extract a set of domain-specific terms $\{\mathcal{J}^{i}\}_{i=1}^{n_{\mathcal{J}}}$:
\begin{equation}
\small
    \{\mathcal{J}^{i}\}_{i=1}^{n_{\mathcal{J}}} = \mathcal{M}_{ext}(p_{ext}(\{e_t^i\}_{i=1}^{n_{e_t}})),
\end{equation}
where $n_{\mathcal{J}}$ is the number of extracted terms, and $p_{ext}$ is the prompt for domain-specific term extraction. 

\noindent \textbf{Domain-specific Annotator}. To ensure alignment between domain-specific and general annotations, an LLM $\mathcal{M}_{spe}$, acting as a domain-specific annotator, applies the extracted terms $\{\mathcal{J}^{i}\}_{i=1}^{n_{\mathcal{J}}}$ to general annotations $e_g$, converting them into target-domain annotations $e_t$. Formally, this is represented as:
\begin{equation}
\small
    e_t = \mathcal{M}_{spe}(p_{spe}(e_g, \{\mathcal{J}^{i}\}_{i=1}^{n_{\mathcal{J}}})),
\end{equation}
where $p_{spe}$ is the prompt for generating domain-specific annotations.

\noindent \textbf{Annotation Reviewer}. To improve the quality of domain-specific annotations and ensure better alignment with general annotations, we introduce an annotation reviewer. This LLM, $\mathcal{M}_{rev}$, reviews the generated annotations and extracted terms, providing feedback $e_f$ to the extractor and annotator:
\begin{equation}
\small
    e_f = \mathcal{M}_{rev}(p_{rev}(e_g, e_t, \{\mathcal{J}^{i}\}_{i=1}^{n_{\mathcal{J}}})),
\end{equation}
where $p_{rev}$ is the prompt for reviewing annotations. 
This feedback loop ensures more precise term extraction and better alignment between general and domain-specific annotations. Based on the feedback, the extractor $\mathcal{M}_{ext}$ refines the extraction process, and the annotator $\mathcal{M}_{spe}$ enhances its annotations accordingly.

%% file: 4_experiment.tex
\section{Experiments}
This section presents the experimental results. We first evaluate the \method's annotations in downstream tasks and on a synthetic dataset, then examine domain-specific annotations, and finally assess the contribution of key \method components.

\subsection{Experimental Setup}
\label{sec:experimental_setup}
\noindent \textbf{Dataset}. To evaluate the effectiveness of \method, 
five real-world datasets from distinct domains are considered: Stock, Health, Energy, Environment, Social Good, {Climate and Economy}. Specifically, the stock dataset includes 1,935 US stocks with the recent 6-year data, collected by ourselves. The other four datasets come from the public benchmark Time-MMD~\cite{liu2024time}. In this paper, the Stock and Health datasets serve as the source domains, while 
{the rest five datasets}
are treated as the target domains.
Additionally, we generate a synthetic dataset containing both time series and ground-truth annotations to directly assess the quality of general annotations. More details on these datasets can be found in Appendix~\ref{appendix:dataset_statistics}. 

\noindent \textbf{LLMs}. Our experiments utilize one closed-source model, GPT-4o~\cite{achiam2023gpt} and two open-source models, LLaMA3.1-8B~\cite{llama31modelcard} and Qwen2-7B~\cite{qwen2}.

\subsection{Evaluating General Annotations in Downstream Tasks}
\label{sec:indirect_general_evaluation}

To evaluate the quality of the general annotations, we apply the generated annotations to the multi-modal downstream tasks (\textit{i.e.}, time series forecasting and imputation) by following the experimental setup in Time-MMD~\cite{liu2024time}. As in Fig.~\ref{fig:MMTSFlib_overall_framework}, time series data and textual annotations are processed independently by unimodel TSF models and LLMs with projection layers. The model-specific outputs are then fused through a linear weighting mechanism to generate final predictions. Incremental RL-based selection in Section~\ref{sec:adaptive_feature_selection} is used in \method to select the top-$k$ most important features for generating annotations.
The implementation details are provided in Appendix~\ref{appendix:framework_for_multi_modal_downstream_tasks} and~\ref{appendix:implementation_details}.


\noindent \textbf{Baselines}. 
\method is, to the best of our knowledge, the first work on cross-domain multi-modal time series annotation. We compare it with several representative single-domain methods: No-Text, Time-MMD~\cite{liu2024time}, and DirectLLM (which directly uses LLM-generated annotations). Details of these methods are provided in Appendix~\ref{appendix:compared_methods_downstream}.



\noindent \textbf{Evaluation Metrics}. For  time series forecasting task, we use MSE (Mean Squared Error) and MAE (Mean Absolute Error) as evaluation metrics, where lower values for both MSE and MAE mean better annotations. 


\noindent \textbf{Experimental Results}. Table~\ref{tab:general_annotation_evaluation_forecasting} presents the comparison results for the time series forecasting task, where Informer~\cite{zhou2021informer} is the forecasting model and GPT-4o~\cite{achiam2023gpt} serves as the LLM backbone. Additional forecasting results using different LLM backbones are available in Appendix~\ref{appendix:time_series_forecasting_more_resuilts}. The following observations can be made: (1) No-Text shows the worst performance across all datasets, validating the need for annotations to improve performance in downstream tasks. This suggests that better downstream task performance indicates higher-quality annotations. (2) \method achieves the best performance among all compared methods, demonstrating its effectiveness in generating high-quality general annotations. 
Additional results of time series imputation tasks, can be found in Appendix~\ref{appendix:time_series_imputation_more_resuilts}. 


\begin{table}[t]
\centering
\small
\caption{Forecasting results with GPT-4o as the LLM backbone. NT, TM, and DL refer to No-Text, Time-MMD, and DirectLLM, respectively. MSE is shown in the top half and MAE in the bottom half.}
\begin{tabularx}{0.95\linewidth}{l|cccc}
\toprule
\textbf{Domain} & \textbf{NT} & \textbf{TM} & \textbf{DL} & \textbf{TESSA} \\ \midrule
\multirow{1}{*}{Environment}  & 1.2542
& 0.8483 & 0.7714 & \textbf{0.4629} \\
\multirow{1}{*}{Energy}   & 2.0117 & 0.2172 & 0.0575 & \textbf{0.0482} \\
\multirow{1}{*}{Social Good}   & 2.1457
& 1.6072 & 0.4639 & \textbf{0.1935}\\
\midrule
\multirow{1}{*}{Environment}  &  0.7387& 0.6865 & 0.6604 & \textbf{0.4424}  \\
\multirow{1}{*}{Energy}   & 1.1663 & 0.2139 & 0.0055 & \textbf{0.0040} \\
\multirow{1}{*}{Social Good}   & 1.1205 & 0.9731 & 0.3801 & \textbf{0.0825}\\
\bottomrule
\end{tabularx}
\label{tab:general_annotation_evaluation_forecasting}
\end{table}

\subsection{Evaluating General Annotations in Synthetic Datasets} 
\label{sec:direct_general_evaluation}
We construct a synthetic dataset with time series data and ground-truth annotations to validate \method's performance. Implementation details are provided in Appendix~\ref{appendix:general_annotation_synthetic_implementation_details}.

\noindent \textbf{Evaluation Metrics}. 
We apply the LLM-as-a-judge approach~\cite{bubeck2023sparks,dubois2024alpacafarm}, evaluating two metrics: Clarity and Comprehensiveness. Two distinct LLMs score the generated annotations on a scale of $1$ to $5$ for each metric, with an overall score calculated as the mean of the two metrics. 
Further details on the metrics and the LLM-judge prompts can be found in 
Appendix~\ref{appendix:general_annotation_synthetic_implementation_details}.

\noindent \textbf{Experimental Results}. 
We compare \method with DirectLLM in Table~\ref{tab:general_annotation_evaluation_synthetic}. The ``Mean'' denotes the average score of generated annotations for each method, and $\textbf{P(T>D)}$ is the percentage of \method's annotations that receive higher scores than DirectLLM's. The results show that \method outperforms DirectLLM on both metrics, with average scores of $3.90$ in Clarity and $4.44$ in Comprehensiveness, compared to DirectLLM's $3.79$ and $1.55$. Additionally, $82.71\%$ of \method's annotations receive higher scores, indicating that \method produces more essential and easily understandable features, further demonstrating its effectiveness. 

\begin{table}[t]
\centering
\small
\caption{General annotation results on the synthetic dataset with GPT-4o as the LLM backbone.}
\begin{tabularx}{0.9\linewidth}{llcc}
\toprule
\textbf{Metric} & \textbf{Method} & \textbf{Mean} & \textbf{P(T>D) (\%)} \\ \midrule
\multirow{2}{*}{Clarity}  & TESSA & \textbf{3.90} & \multirow{2}{*}{\textbf{69.76}}
\\
  & DirectLLM & 3.79 & \\ 
\midrule
\multirow{2}{*}{Compre.}  & TESSA & \textbf{4.44} & \multirow{2}{*}{\textbf{87.10}}
\\
  & DirectLLM & 1.55 & \\ 
  \midrule
\multirow{2}{*}{Overall}  & TESSA & \textbf{4.14}& \multirow{2}{*}{\textbf{82.71}}
\\
  & DirectLLM & 2.84 & \\ 

\bottomrule
\end{tabularx}
\label{tab:general_annotation_evaluation_synthetic}
\end{table}

\subsection{Domain-specific Annotation Evaluation}
\label{sec:domain_specific_annotation_evaluation}
In this subsection, we evaluate the quality of domain specific annotations. 
Similar to Section~\ref{sec:direct_general_evaluation}, we adopt a LLM-as-a-Judger strategy to evaluate the performance of domain-specific annotation agent from three perspectives: Clarity, Comprehensiveness, and Domain-relevance. The overall score is the average of these three metrics. Further details on these metrics are provided in Appendix~\ref{appendix:domain_specific_annotation_evaluation_metric}. 

\noindent \textbf{Experimental Results}. 
We present the comparison results of \method and DirectLLM on the Environment dataset in Table~\ref{tab:domain_specific_annotation_evaluation}, with GPT-4o as the LLM backbone. The key observations are: (1) \method significantly outperforms DirectLLM across all metrics, achieving an overall score of $4.64$ compared to DirectLLM's $3.41$. Notably, $98.51\%$ of \method's annotations receive higher scores, demonstrating its effectiveness in generating high-quality domain-specific annotations. (2) \method scores $4.74$ in Clarity and $4.38$ in Comprehensiveness, while DirectLLM scores $3.32$ and $3.01$, respectively. This shows that \method’s annotations are clearer, more concise, and cover more important features. (3) \method also excels in domain relevance, with $94.72\%$ of its annotations scoring higher, achieving an average of $4.30$, significantly outperforming DirectLLM's $3.41$. This indicates that \method produces highly accurate annotations that effectively use domain-specific terminology and maintain strong contextual relevance. More results on other datasets are in Appendix~\ref{appendix:domain_specific_evaluation_more_results}.


\begin{table}[t]
\centering
\small
\caption{Domain-specific annotation results on the Environment dataset using GPT-4o as the LLM. Dom. Rel. is the domain-relevance metric used in Section~\ref{sec:domain_specific_annotation_evaluation}.}

\begin{tabularx}{0.9\linewidth}{llcc}
\toprule
\textbf{Metric} & \textbf{Method} & \textbf{Mean} & \textbf{P(T>D) (\%)} \\ \midrule
\multirow{2}{*}{Clarity}  & TESSA & \textbf{4.74} & \multirow{2}{*}{\textbf{99.81}}
\\
  & DirectLLM & 3.32 & \\ 
\midrule
\multirow{2}{*}{Compre.}  & TESSA & \textbf{4.38} & \multirow{2}{*}{\textbf{97.04}}
\\
  & DirectLLM & 3.01 & \\ 
  \midrule
\multirow{2}{*}{Dom. Rel.}  & TESSA & \textbf{4.30} & \multirow{2}{*}{\textbf{94.72}}
\\
  & DirectLLM & 3.57 & \\ 
  \midrule
\multirow{2}{*}{Overall}  & TESSA & \textbf{4.64 }& \multirow{2}{*}{\textbf{98.51}}
\\
  & DirectLLM & 3.41 & \\ 

\bottomrule
\end{tabularx}
\label{tab:domain_specific_annotation_evaluation}
\end{table}

\nop{\subsection{In-depth Dissection of TESSA}
\label{sec:in_depth_TESSA}
\noindent\textbf{Comparison between Adaptive Feature Selections}.
We conduct experiments to compare our two adaptive feature selection methods, offline selection and incremental selection. 
To evaluate the effectiveness of two appraoches in selecting top-$k$ most important features, we evaluate the qualities of the generated general annotations and domain-specific annotations by applying the two methods. We follow Section~\ref{sec:indirect_general_evaluation} and Section~\ref{sec:domain_specific_annotation_evaluation} to evaluate their qualities, respectively. 
Environment is set as the target domain, and the results are shown in Fig.~\ref{fig:comparison_llm_rl}. From the results, we observe that TESSA achieves comparable performance in both general annotation generation and domain-specific annotation generation when using the two feature selection methods. 
Specifically, according to Fig.~\ref{fig:comparison_llm_rl}(a), in general annotation evaluation, both approaches achieve MSE and MAE around $0.46$ and $0.44$, respectively. And as the domain-specific annotation evaluation shown in Fig.~\ref{fig:comparison_llm_rl}(b), both approaches achieve consistently high scores across the three metrics, further indicating their similar effectiveness in selecting the top-$k$ most important features. However, incremental selection is more cost-effective as it reduces redundant re-querying of previously used data.}

\subsection{In-depth Dissection of TESSA}
\label{sec:in_depth_TESSA}
\begin{figure}[t]
    \small
    \centering
    \begin{subfigure}{0.45\linewidth}   
    \label{fig:comparison_llm_rl_general}
        \includegraphics[width=0.98\linewidth]{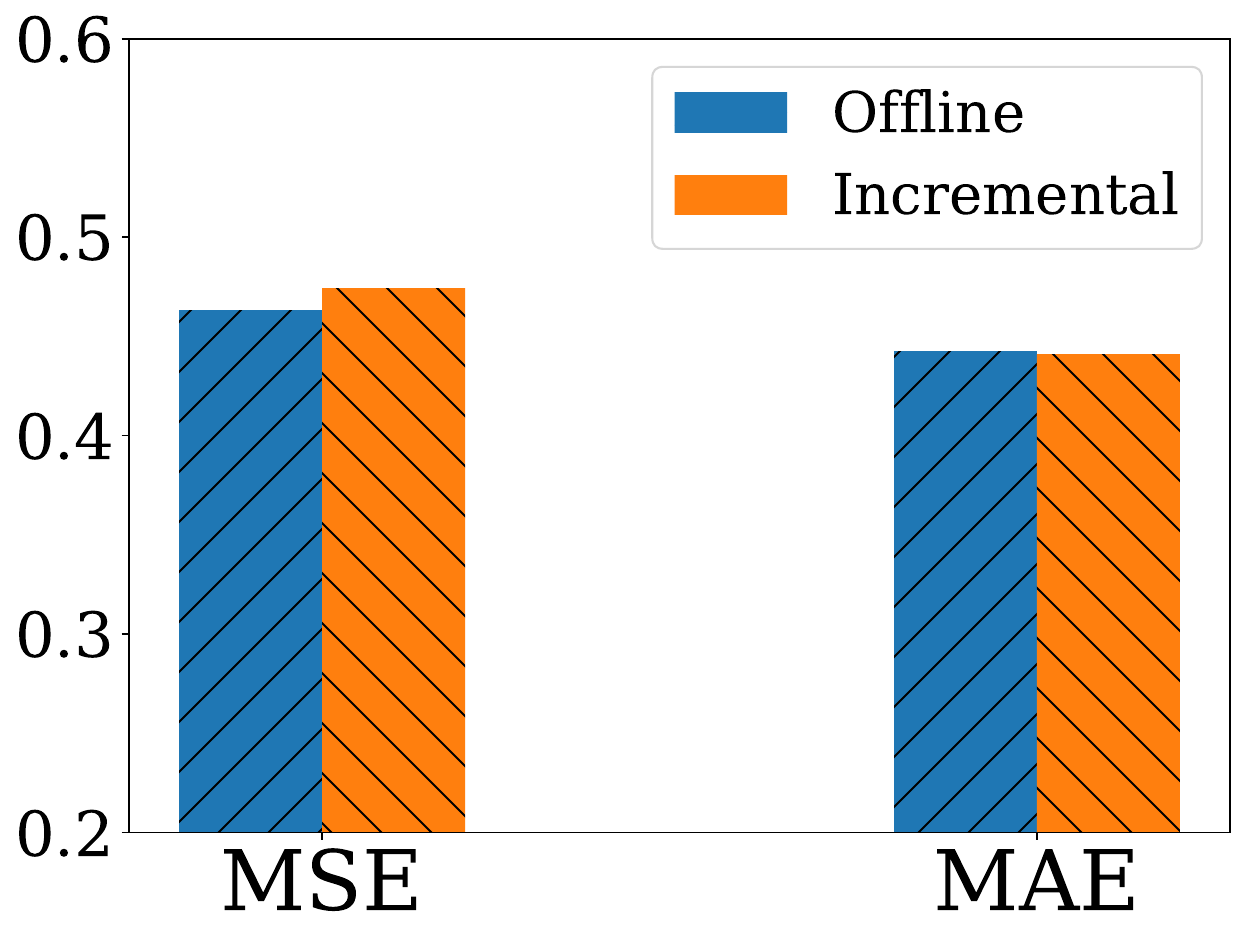}
       \vskip -0.5em
        \caption{General}
    \end{subfigure}
    \begin{subfigure}{0.45\linewidth}
    \label{fig:comparison_llm_rl_specific}
        \includegraphics[width=0.98\linewidth]{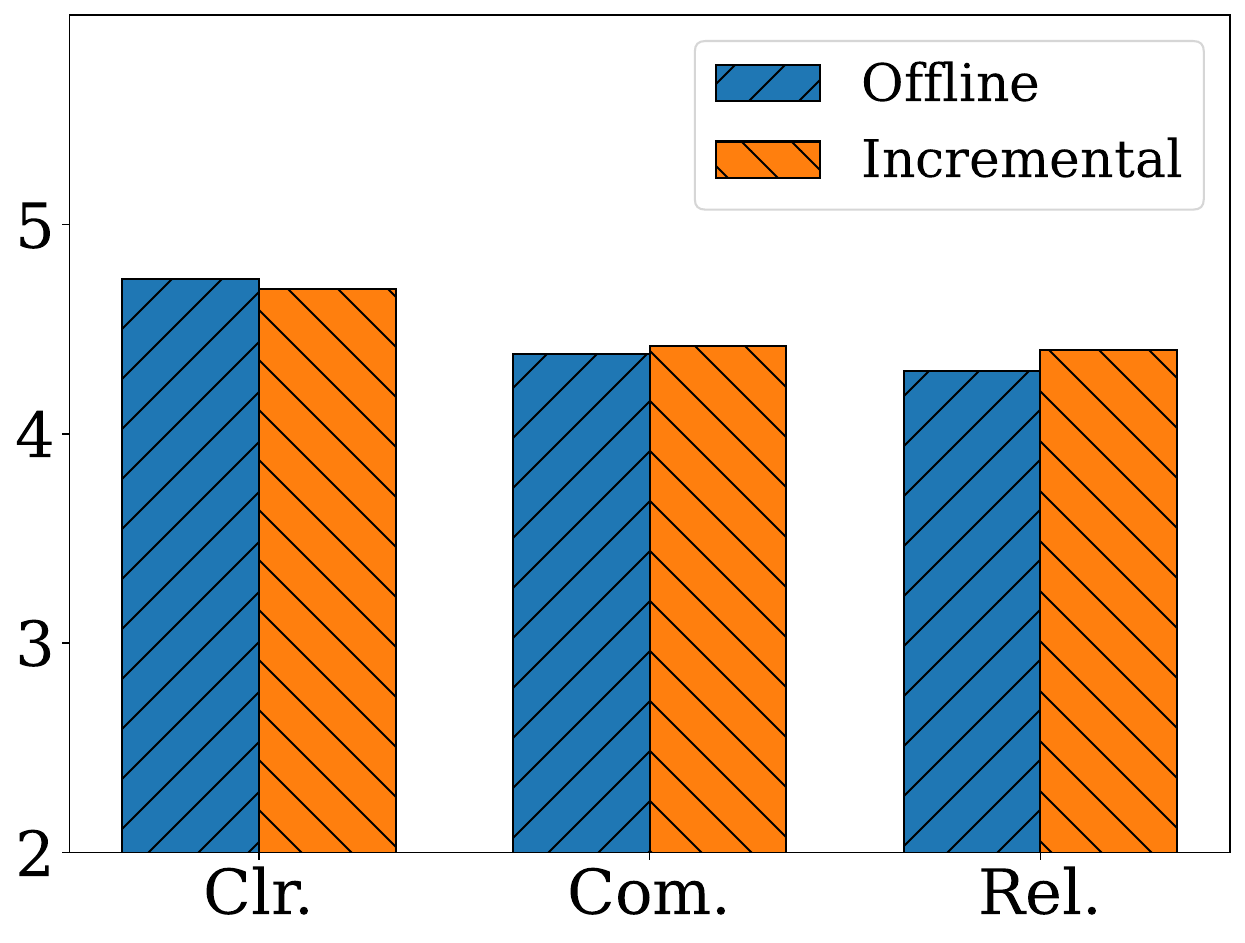}
       \vskip -0.5em
        \caption{Specific}
 
    \end{subfigure}
   \vskip -1.0em
\caption{Comparison of offline vs. incremental feature selection. GPT-4o is the LLM backbone, with Environment as the target domain. (a) General annotation results; (b) Domain-specific annotation results.}
    \label{fig:comparison_llm_rl}
\end{figure}

\noindent\textbf{Adaptive Feature Selections}. We compare our two feature selection methods: offline LLM-based selection and incremental RL-based selection. To assess their effectiveness in selecting the top-$k$ most important features, we evaluate the quality of the generated general and domain-specific annotations, following the procedures in Sections~\ref{sec:indirect_general_evaluation} and~\ref{sec:domain_specific_annotation_evaluation}. Environment is set as the target domain, with results shown in Fig.~\ref{fig:comparison_llm_rl}. The results indicate that \method performs comparably in both general and domain-specific annotation generation using either selection method. Specifically, as shown in Fig.~\ref{fig:comparison_llm_rl}(a), both approaches achieve MSE and MAE around $0.46$ and $0.44$ for general annotations. Similarly, in Fig.~\ref{fig:comparison_llm_rl}(b), both methods score consistently high across all domain-specific metrics, demonstrating their effectiveness in selecting important features. However, incremental RL-based selection proves more cost-effective by reducing redundant re-querying of previously used data.


\noindent\textbf{Ablation Studies}. We perform ablation studies to assess the importance of domain decontextualization and adaptive feature selection in \method. To evaluate domain decontextualization, we introduce a variant, \method/D, which bypasses the domain decontextualization LLM and directly extracts text-wise features from domain-specific annotations. Table~\ref{tab:ablation_study_extracted_features} shows that \method/D captures irrelevant features, such as \textit{higher prices over time} and \textit{fun}, which are unrelated to time series analysis. This supports our claim that domain-specific terminology can hinder the accurate extraction of time-series-relevant features.

To prove the importance of adaptive feature selection in \method, we remove the adaptive feature selection module to create a variant, \method/F. We apply an LLM-as-a-judger to compare the quality of the generated annotations between \method and its variants. The evaluation metrics are introduced in Appendix~\ref{appendix:ablation_study_evaluation_metric}. We choose Social Good as the target dataset. The comparison results are presented in Table~\ref{tab:ablation_studies}, with qualitative examples provided in Appendix~\ref{appendix:ablation_study_qualitative_examples}. We observe that: \method consistently outperforms \method/F. Specifically, \method achieves a clarity score of $4.41$, compared to $3.66$ for \method/F. This demonstrates the necessity of adaptive feature selection. Furthermore, according to Table~\ref{tab:ablation_study_impact_of_feature_selection} in Appendix~\ref{appendix:ablation_study_qualitative_examples}, the annotations generated by \method/F tend to include many features without proper analysis. This shows that involving too many features can hinder the clarity of the annotations, further emphasizing the importance of adaptive feature selection in improving annotation quality. 
Additional ablation studies examining the contributions of other components of \method are in Appendix~\ref{appendix:more_ablation_studies_tessa_components}. And discussions on data contamination are provided in Appendix~\ref{appendix:ablation_studies_fully_open_LLMs}.

\begin{table}[t]
\centering
\small
\caption{Ablation studies in the SocialGood dataset. GPT-4o is the LLM backbone.}
\begin{tabularx}{0.85\linewidth}{llcc}
\toprule
\textbf{Metric} & \textbf{Method} & \textbf{Mean} & \textbf{P(T>D) (\%)} \\ \midrule
\multirow{2}{*}{Clarity}  & TESSA & \textbf{4.41} & \multirow{2}{*}{\textbf{83.3}}
\\
  & TESSA/F & 3.66 & \\ 

\bottomrule
\end{tabularx}
\label{tab:ablation_studies}
\end{table}

\subsection{Case Study of TESSA}
\label{sec:case_study_main}

We conduct a case study to further validate the effectiveness of \method. A representative time series from the Social Good domain (Fig.~\ref{fig:case_study_ts_data}(b)) is selected, and both \method and DirectLLM are applied to generate general and domain-specific annotations, summarized in Table~\ref{tab:case_study_main}. To assess the quality of the annotations, we use an LLM-as-a-judger to evaluate the domain-specific annotations from both methods, with results shown in Table~\ref{tab:case_study_domain_specific_evaluation_main}. Our findings indicate that: (1) \method's general annotations capture more meaningful patterns, aiding user understanding and downstream tasks, whereas DirectLLM only highlights basic trends; and (2) \method's domain-specific annotations consistently outperform DirectLLM across all metrics, offering clearer, more comprehensive, and contextually relevant insights.
More case studies of multivariate time series data are provided in Appendix~\ref{appendix:case_study}.


%% file: 6_conclusion.tex
\section{Conclusion}

In this work, we introduce \method, a multi-agent system for automatic general and domain-specific time series annotation. \method incorporates two agents, a general annotation agent and a domain-specific annotation agent, to extract and leverage both time-series-wise and text-wise knowledge from multiple domains for annotations. \method overcomes the limitations of directly applying LLMs, which often capture only basic patterns and may hallucinate, by effectively identifying and emphasizing significant patterns in time series data. Our experiments on synthetic and real-world datasets from diverse domains demonstrate the effectiveness of \method in generating high-quality general and domain-specific annotations.

\nop{\section{Limitations}
Potential limitations of this work include the need for limited target-domain annotations to learn the domain-specific jargon for automatic domain-specific annotation generation. Additionally, our approach requires annotations from source domains to transfer knowledge to target domains. }

\section{Limitations}
Potential limitations of this work include the need for limited target-domain annotations to learn domain-specific jargon for generating domain-specific annotations. Additionally, our approach relies on annotations from source domains to transfer knowledge to target domains. If the chosen source domain annotations are all of low quality or lack sufficient common knowledge, it may affect the overall performance of \method.

\section{Ethics Statement}
We adhere to the ACM Code of Ethics in our research. All datasets and models used in this study are either publicly accessible or synthetically generated. Specifically, we created a synthetic dataset comprising time series data with generated general annotations to facilitate our experiments while avoiding the use of any personal or sensitive real-world data. 
We acknowledge the potential risks and harms associated with LLMs, such as generating harmful, offensive, or biased content. Moreover, LLMs are often prone to generating incorrect information, sometimes referred to as hallucinations. We recognize that the models studied in this paper are not exceptions to these limitations. Previous research has shown that the LLMs used in this study suffer from bias, hallucinations, and other issues. We emphasize the importance of responsible and ethical use of LLMs and the need for further research to mitigate these challenges before deploying them in real-world applications. The models used in this work are licensed under the terms of OpenAI, LLaMA, and Qwen.


%% file: 7_appendix.tex
\section{More Related Work}
\label{appendix:related_works}

\noindent\textbf{LLMs for Time Series Analysis}.
\label{appendix:related_works_llm_TSA}
The rapid advancement of LLMs in natural language processing has unveiled unprecedented capabilities in sequential modeling and pattern recognition, which can be leveraged for time series analysis. Three primary approaches are commonly adopted~\cite{jiang2024empowering}: direct querying of LLMs~\cite{xue2023promptcast,yu2023temporal,gruver2024large}, fine-tuning LLMs with task-specific modifications~\cite{chang2023llm4ts,cao2024tempo,jin2024time,sun2024test}, and incorporating LLMs into time series models to enhance feature extraction~\cite{li2024frozen}.

Direct querying involves using LLMs to generate predictions or identify patterns from the data without modifying the underlying architecture. For example, PromptCast~\cite{xue2023promptcast} applies LLMs to time series forecasting through a sentence-to-sentence paradigm. \citeauthor{yu2023temporal} explore the use of LLMs for domain-specific tasks like financial time series forecasting~\cite{yu2023temporal}, while LLMTime~\cite{gruver2024large} demonstrates how LLMs can function as effective learners by tokenizing time series data in a text-like format.

Fine-tuning LLMs enables them to better capture the intricacies of time series data by adapting them to specific datasets or tasks. For instance, LLM4TS~\cite{chang2023llm4ts} shows that fine-tuning pre-trained models can enhance forecasting performance. Additionally, TEMPO~\cite{cao2024tempo} and TEST~\cite{sun2024test} introduce architectures tailored for time series prediction, further demonstrating the power of specialized designs.

Lastly, LLMs can also act as feature enhancers within traditional time series models, enriching data representations and boosting performance. For example, \cite{li2024frozen} illustrates how a frozen LLM can augment zero-shot learning for ECG time series analysis, highlighting the potential of LLMs to provide valuable features for complex datasets.

\noindent\textbf{Domain Specialization of LLMs}. 
Domain specialization of LLMs refers to the process of adapting broadly trained models to achieve optimal performance within a specific domain. This is generally categorized into three approaches: prompt crafting~\cite{ben2022pada,zhang2023automatic,xu2024llm}, external augmentation~\cite{izacard2023atlas}, and model fine-tuning~\cite{malik2023udapter,pfeiffer2020adapterfusion}. One of the earliest efforts in this area is PADA~\cite{ben2022pada}, which enhances LLMs for unseen domains by generating domain-specific features from test queries and using them as prompts for task prediction. Auto-CoT~\cite{zhang2023automatic} advances domain specialization by prompting LLMs with the phrase ``Let’s think step by step,'' helping guide the models in generating reasoning chains. Additionally, \citet{izacard2023atlas} propose integrating a relatively lightweight LLM with an external knowledge base, achieving performance comparable to much larger models like PaLM~\cite{chowdhery2023palm}. These studies highlight the flexibility of LLMs in adapting to specific domains through various strategies for domain adaptation.

\section{Notations}
\label{appendix:notations}
Table~\ref{tab:notation} presents all the notations we used in this paper.
\begin{table}[ht]
\centering
\small
\caption{Notation Table}
\begin{tabularx}{1\linewidth}{cp{0.70\linewidth}}
\toprule
\textbf{Symbol} & \textbf{Description} \\ \midrule
$\mathbf{x}$ & Input time series data \\ 
$e_s$ & Domain-specific annotation from source domains  \\ 
$e_t$ & Domain-specific annotation from target domain  \\ 
$e_d$ & Domain-decontextualized annotation \\ 
$e_g$ & General annotation \\ 
$f_t$ & Time-series-wise feature \\ 
$f_l$ & Text-wise feature \\ 
$\mathcal{J}$ & Domain-specific term (jargon) from target domain \\ \midrule
$\mathcal{M}_d$ & Domain decontextualizer \\ 
$\mathcal{M}_t$ & Time-series-wise feature extractor \\ 
$\mathcal{M}_l$ & Text-wise feature extractor \\ 
$\mathcal{M}_{sel}$ & Feature selector \\ 
$\mathcal{M}_{gen}$ & General annotator \\ 
$\mathcal{M}_{jar}$ & domain-specific term extractor \\ 
$\mathcal{M}_{spe}$ & Domain-specific annotator \\ 
$\mathcal{M}_{rev}$ & Annotation reviewer \\ \midrule
$p_{de}$ & prompt of domain-decontextualization\\ 
$p_{l}$ & prompt of text-wise feature extraction\\ 
$p_{score}$ & prompt of scoring\\ 
$p_{gen}$ & prompt of general annotation\\ 
$p_{ext}$ & prompt of domain-specific term extraction \\ 
$p_{spe}$ & prompt of domain-specific annotation \\ 
$p_{rev}$ & prompt of annotation review \\ 

\bottomrule
\end{tabularx}
\label{tab:notation}
\end{table}

Additionally, we also provide some specific examples of domains, annotations, and features to improve the clarity of the problem settings of cross-domain time series annotation defined in Section~\ref{sec:methdology}. Specifically, our paper consider six distinct domains, i.e., {Stock, Health, Environment, Social Good, Climate and Economy}. For instance, the stock dataset includes multivariate time series data (e.g., {stock price, volume, RSI, moving average}) with corresponding {annotations capturing features like support levels and resilience}. More examples of text-wise features can be found in Table~\ref{tab:ablation_study_extracted_features}. More examples of domain-specific annotations are provided in Tables~\ref{tab:case_study_environment_more_1},~\ref{tab:case_study_environment_more_2},~\ref{tab:case_study_energy_more_1},~\ref{tab:case_study_energy_more_2},~\ref{tab:case_study_socialgood_more_1} and~\ref{tab:case_study_socialgood_more_2} of Appendix~\ref{appendix:case_study}.`

\section{More Details of Multi-modal Feature Extraction}
\label{appendix:more_details_time_series_feature_extraction}
\subsection{Time-series Feature Extraction}
\label{appendix:time_series_wise_extraction}
\nop{Given a time series data $\mathbf{X}=\{(\mathbf{x}_1,\cdots,\mathbf{x}_L)\}$, we develop a time series extraction toolbox $\{f_t^1,\ldots,f_t^{N_t}\}$ to extract time-series-wise features from $\mathbf{X}$. Specifically, we include \textit{seasonality}, \textit{trend}, \textit{noise}, \textit{moving average}, \textit{lag feature}, \textit{rolling window feature} and \textit{Fourier frequency} as
intra-variable time-series-wise features. For multivariate time series, we also consider the inter-variable time-series-wise features, \textit{i.e.}, \textit{mutual information}, \textit{pearson correlation} and \textit{canonical correlation}.

Especially, we employ Seasonal-Trend decomposition (STL)~\cite{cleveland1990stl} to extract seasonality, trend and noise from the given time series data. To extract Fourier frequencies, the Fast Fourier Transform (FFT)~\cite{almeida1994fractional} is applied to convert a time-domain signal into its frequency components. For the inter-variable time-series features, we use \texttt{np.corrcoef} to compute the Pearson correlation. To calculate \textit{mutual information}, two time series are first discretized, followed by 
 \texttt{{sklearn.metrics.mutual\_info\_score}}. To calcualte \textit{canonical correlation}, we first use \texttt{sklearn.cross\_decomposition} to decompose two time series data, and then use \texttt{np.corrcoef} to obtain it.}

 Given a time series data $\mathbf{X} = \{(\mathbf{x}_1, \cdots, \mathbf{x}_L)\}$, we develop a time series extraction toolbox $\{f_t^1,\ldots,f_t^{N_t}\}$ to extract time-series-wise features from $\mathbf{X}$. Specifically, we include \textit{seasonality}, \textit{trend}, \textit{noise}, \textit{moving average}, \textit{lag feature}, \textit{rolling window feature}, and \textit{Fourier frequency} as intra-variable time-series-wise features. For multivariate time series, we also consider inter-variable time-series-wise features, i.e., \textit{mutual information}, \textit{Pearson correlation}, and \textit{canonical correlation}.

In particular, we employ Seasonal-Trend decomposition (STL)~\cite{cleveland1990stl} to extract seasonality, trend, and noise from the given time series data. To extract Fourier frequencies, the Fast Fourier Transform (FFT)~\cite{almeida1994fractional} is applied to convert a time-domain signal into its frequency components. For the inter-variable time-series features, we use \texttt{np.corrcoef} to compute the Pearson correlation. To calculate \textit{mutual information}, two time series are first discretized, followed by \texttt{sklearn.metrics.mutual\_info\_score}. To calculate \textit{canonical correlation}, we first use \texttt{sklearn.cross\_decomposition} to decompose two time series data, and then use \texttt{np.corrcoef} to obtain the correlation.




\section{More Details of Adaptive Feature Selection}
\subsection{Offline LLM-based Feature Selection}
\label{appendix:more_details_llm_based_feature_selection}
\nop{The templates for $p_{score}$ in Eq.~(\ref{eq:score_llm_select}) are shown in Fig.~\ref{fig:prompt_time_series_feature_score} and Fig.~\ref{fig:prompt_text_feature_score}, respectively. 

In some cases, we cannot input all the annotations to LLMs for calculating scores. We may split the annotations into several small batchs and input the annotations in the small batches to calculate the score using Eq.~(\ref{eq:score_llm_select}).
After that, we will then accumulate the scores in all batches to get the final scores of each feature and then select the features with top-$k$ highest scores.}


In some cases, we cannot input all the annotations to LLMs for calculating scores. We may split the annotations into several small batches and input the annotations in the small batches to calculate the score using Eq.~(\ref{eq:score_llm_select}). After that, we will accumulate the scores from all batches to get the final scores of each feature/token and then select the features with the top-$k$ highest scores.

\subsection{Incremental Reinforcement Learning-based Feature Selection}
\label{sec:discussion_rl_selection}

\noindent \textbf{The necessity of this component}. In the proposed LLM-based feature selection from Section~\ref{sec:methdology}, when new annotations exhibit different distributions or feature characteristics compared to the old data, it becomes necessary to re-query both old and new data to select the top-$k$ most important features. This process is computationally intensive and resource-inefficient, especially as the volume of data grows. To address this issue, we propose the incremental reinforcement learning (RL)-based feature selection method. This approach provides the following benefits:
\begin{itemize}[leftmargin=*]
    \item \textbf{Cost-Efficiency}: Instead of re-querying LLMs with all the data, the RL-based method incrementally updates the knowledge stored in policy networks, requiring only the new data during updates.
    \item \textbf{Scalability}: By reducing redundant computations, the incremental RL-based method ensures scalability in dynamic environments with evolving data.
\end{itemize}

\section{Time Complexity Analysis}
To analyze the time complexity of \method, we consider each component of \method separately, focusing on the computational cost associated with feature extraction, feature selection, and annotation generation and review.

\noindent\textbf{Feature Extraction}. Extracting intra-variable and inter-variable features has a complexity of  $\mathcal{O}(C^2 \cdot n)$ , where $C$ is the number of channels in the time series and $n$ is the number of time points.

\noindent\textbf{Feature Selection}. For offline LLM-based feature selection, the complexity is $\mathcal{O}(k \cdot M \cdot L^2)$, where $k$ is the number of features, $M$ is the model size (number of parameters) of the LLM, and $L$ is the input sequence length. Incremental RL-based selection reduces this overhead by incrementally updating policy networks without re-querying old data.

\noindent\textbf{Annotation Generation and Review}. Each LLM inference for annotation generation or review has a complexity of $\mathcal{O}(M \cdot L^2)$. Given $T$ samples, the overall complexity becomes $\mathcal{O}(T \cdot M \cdot L^2)$.

\noindent\textbf{Overall Complexity}. The combined complexity can be expressed as:
\begin{equation*}
     \mathcal{O}(T \cdot [C^2 \cdot n + k \cdot M \cdot L^2]),
\end{equation*}
where $T$ is the number of time series samples.

\section{Experimental Settings}
\subsection{Dataset Statistics}
\label{appendix:dataset_statistics}
\textbf{Datasets}. To evaluate the effectiveness of \method, 
five real-world datasets from distinct domains are considered: Stock, Health, Energy, Environment, and Social Good. Specifically, the stock dataset includes 1,935 US stocks with the recent 6-year data, collected from Investtech\footnote{https://www.investtech.com/}. The other four datasets come from the public benchmark Time-MMD~\cite{liu2024time}. 
The dataset statistics are summarized in Table~\ref{tab:dataset_statistic}.

Additionally, we generate a synthetic dataset containing both time series and ground-truth annotations to directly assess the quality of the general annotations. The synthetic dataset is created by combining several key components from the time-series data:
\begin{itemize}[leftmargin=*]
    \item \textbf{Trend}: Introduces an overall direction, which can be upward, downward, or mixed.
    \item \textbf{Seasonality}: Adds cyclical patterns, modeled using sine waves.
    \item \textbf{Fourier Feature}: Incorporates complex periodic behavior by combining multiple sine and cosine waves.
    \item \textbf{Noise}: Adds Gaussian noise to simulate random fluctuations and real-world imperfections.
    \item \textbf{Rolling Window Features}: Captures smoothed trends (mean) and local variability (max/min).
    \item \textbf{Lag Features}: Uses past values to capture autocorrelation in the time series.
\end{itemize}
Ground-truth annotations are then generated by summarizing the key components of the synthetic time series.

\nop{To generate such a dataset, from time-series-wise, the synthetic time series dataset is generated by combining several components:
\begin{itemize}[leftmargin=*]
    \item Trend: Adds an overall direction (upward, downward, mixed).
    \item Seasonality: Introduces cyclical patterns using a sine wave.
    \item Fourier Feature: Adds complex periodic behavior by combining multiple sine and cosine waves.
    \item Noise: Gaussian noise is added for random fluctuations, mimicking real-world imperfections.
    \item Rolling Window Features: Provides smoothed trend (mean) and local variability (max/min).
    \item Lag Features: Adds past values as features to capture autocorrelation.
\end{itemize}
We can then generate ground-truth annotations by summarizing the key components of synthetic time series.} 

In our synthetic dataset, we conduct $100$ times random generation of each components and then combine them together to get $100$ synthetic time series data, each with corresponding textual annotation.

\begin{table*}[t]
\centering
\caption{Dataset Statistics}
\begin{tabularx}{0.75\linewidth}{l|lccc}
\toprule
\textbf{Domain} & \textbf{Frequency} & \textbf{\# Channels} & \textbf{\# Timestamps} & \textbf{\# Samples}  \\ \midrule
Stock & Daily & 4 & 854,878 & 1,758\\
Health & Weekly & 1 & 1,389 & 1,356\\
Social Good & Monthly & 1 & 916 & 497\\
Energy & Daily & 1  & 1,622 & 1,586 \\ 
Environment & Daily & 1 & 11,102 & 1,935 \\
Climate & Monthly & 5 & 496 & 177\\
Economy & Monthly & 3 & 423 & 410\\

\bottomrule
\end{tabularx}
\label{tab:dataset_statistic}
\end{table*}

\subsection{Baseline Methods}
\label{appendix:compared_methods_downstream}
\nop{Three baselines are applied in our general annotation evaluation in downstream tasks:
\begin{itemize}[leftmargin=*]
    \item \textbf{No text}  that no textual data are utilized in forecasting.
    \item \textbf{Time-MMD}~\cite{liu2024time} is a multimodal benchmark for time series analysis, which provides both time series and text data. To adapt this method to our setting, we directly apply the original text data in target datasets from~\cite{liu2024time} to the forecasting task.
    \item \textbf{DirectLLM} directly use the annotations generated by LLMs for time series forecasting. In this paper, several representative LLMs are considered in comparisons. 
\end{itemize}}

Three baselines are applied in our general annotation evaluation for downstream tasks:
\begin{itemize}[leftmargin=*]
    \item \textbf{No Text}: No textual data are utilized in the forecasting process.
    \item \textbf{Time-MMD}~\cite{liu2024time}: A multimodal benchmark for time series analysis that incorporates both time series and text data. To adapt this method to our setting, we apply the original text data from the target datasets in~\cite{liu2024time} to the forecasting task.
    \item \textbf{DirectLLM}: Directly uses the annotations generated by LLMs for time series forecasting. In this paper, we compare several representative LLMs in our evaluations.
\end{itemize}

\subsection{Framework for Multi-modal Downstream tasks}
\label{appendix:framework_for_multi_modal_downstream_tasks}
\begin{figure}[t]
    \centering
    \includegraphics[width=0.98\linewidth]{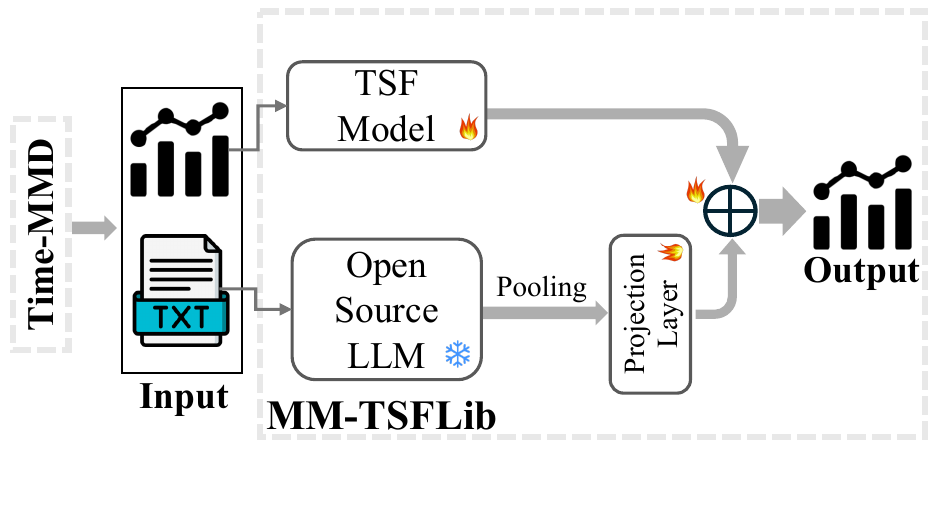}
   \vskip -1.5em
    \caption{Overall framework of MM-TSFlib from Time-MMD~\cite{liu2024time} used in our multi-modal downstream tasks. MMTSFlib uses a model-agnostic multimodal integration framework that independently models time-series and textual annotations within an end-to-end training manner. MM-TSFlib slightly increases the number of trainable parameters, balancing effectiveness and
efficiency.}
    \label{fig:MMTSFlib_overall_framework}
\end{figure}

To evaluate the quality of general annotations, we leverage the multi-modal time series analysis framework proposed in Time-MMD~\cite{liu2024time}, illustrated in Fig.~\ref{fig:MMTSFlib_overall_framework}. Using time series forecasting as a representative task, this framework employs an end-to-end pipeline that combines open-source language models with diverse time-series forecasting (TSF) models. Time-series data and textual annotations are modeled independently through dedicated unimodal TSF architectures and language models (LLMs) equipped with projection layers. The outputs of these modalities are fused via a dynamic linear weighting mechanism to generate final predictions. To optimize computational efficiency, we keep the LLM parameters frozen during training and update only the projection layers. Additionally, pooling layers are introduced to resolve dimension mismatches between textual variables and time-series features. The framework supports end-to-end training with minimal parameter overhead, ensuring both scalability and practicality.

\section{Additional Results for General Annotation Evaluation in Downstream Tasks}
\subsection{Implementation Details}
\label{appendix:implementation_details}
\nop{\noindent\textbf{Time Series Forecasting Models}.
We use Informer~\cite{zhou2021informer} as the forecasting model in the time series forecasting downstream task. We set $\text{dropout}=0.1$, $\text{learning rate}=0.0001$.

\noindent\textbf{Large Language Models}.
We utilize GPT-4o~\cite{achiam2023gpt} and two open-source models, LLaMA3.1-8B~\cite{llama31modelcard} and Qwen2-7B~\cite{qwen2}. For the two open-sourced models, we set \texttt{temperature=1}, \texttt{max\_tokens=2048}. All other settings follow default settings.

In our paper, each experiment is conducted five times and reports the average result. All models are trained on a Nvidia A6000 GPU with 48GB of memory.}

\noindent\textbf{Time Series Forecasting Models}.
We use Informer~\cite{zhou2021informer} as the forecasting model for the time series forecasting task. The model is configured with a dropout rate of $0.1$ and a learning rate of $0.0001$.

\noindent\textbf{Large Language Models}.
We utilize GPT-4o~\cite{achiam2023gpt}, along with two open-source models: LLaMA3.1-8B~\cite{llama31modelcard} and Qwen2-7B~\cite{qwen2}. For the open-source models, we set \texttt{temperature=1} and \texttt{max\_tokens=2048}, while all other settings follow the defaults.

Each experiment in our paper is conducted five times, with the average result reported. All models are trained on an Nvidia A6000 GPU with 48GB of memory.

\subsection{Evaluation in Time Series Forecasting Tasks}
\label{appendix:time_series_forecasting_more_resuilts}

The full results are presented in Table~\ref{tab:general_annotation_evaluation_forecasting_full}. From the table, we can observe the following: (1) \method consistently outperforms all baselines across all settings, demonstrating its effectiveness in generating high-quality general annotations. (2) Among the three LLMs, GPT-4o-backed \method achieves the best performance, outperforming both LLaMA3.1-8B and Qwen2-7B. We attribute this to the higher quality of the annotations generated by GPT-4o compared to the other models, further emphasizing that high-quality annotations can significantly enhance downstream task performance.

\begin{table*}[ht]
\centering
\small
\caption{Comparison results in forecasting. Informer is the time series forecasting model. }
\begin{tabularx}{0.8\linewidth}{lllcccc}
\toprule
\textbf{Domain} & \textbf{Backbone} & \textbf{Metrics} & \textbf{No Text} & \textbf{Time-MMD} & \textbf{DirectLLM} & \textbf{TESSA} \\ \midrule
\multirow{6}{*}{Environment}  & \multirow{2}{*}{GPT-4o} & MSE & 1.2542
& 0.8483 & 0.7714 & \textbf{0.4629} \\
  &  & MAE & 0.7387& 0.6865 & 0.6604 & \textbf{0.4424}  \\ 
  \cmidrule{3-7}
    &  \multirow{2}{*}{LLaMA3.1-8B} & MSE & 1.2542
& 0.8483 & 0.8108 & \textbf{0.5654}   \\ 
      &  & MAE &  0.7387& 0.6865 & 0.6805 &  \textbf{0.5128}   \\ 
    \cmidrule{3-7}
    &  \multirow{2}{*}{Qwen2-7B} & MSE & 1.2542
& 0.8483 & 0.7956 & \textbf{0.5824}   \\ 
      &  & MAE &  0.7387& 0.6865 & 0.6729 & \textbf{0.5419}  \\
\midrule
\multirow{6}{*}{Energy}  & \multirow{2}{*}{GPT-4o} & MSE & 2.0117 & 0.2172 & 0.0575 & \textbf{0.0482} \\
  &  & MAE & 1.1663 & 0.2139 & 0.0055 & \textbf{0.0040}   \\ 
    \cmidrule{3-7}
    &  \multirow{2}{*}{LLaMA3.1-8B} & MSE &  2.0117 & 0.2172  & 0.1023 & \textbf{0.0531}   \\ 
      &  & MAE &  1.1663 & 0.2139  & 0.0130 & \textbf{0.0049}   \\ 
    \cmidrule{3-7}
    &  \multirow{2}{*}{Qwen2-7B} & MSE &  2.0117 & 0.2172  & 0.0824 & \textbf{0.0522}   \\ 
      &  & MAE &  1.1663 & 0.2139  & 0.0097 & \textbf{0.0048}   \\ 
\midrule
\multirow{6}{*}{Social Good}  & \multirow{2}{*}{GPT-4o} & MSE & 2.1457
& 1.6072 & 0.4639 & \textbf{0.1935}\\
  &  & MAE & 1.1205 & 0.9731 & 0.3801 & \textbf{0.0825}\\ 
  \cmidrule{3-7}
    &  \multirow{2}{*}{LLaMA3.1-8B} & MSE & 2.1457 & 1.6072 & 0.6720 &  \textbf{0.3422}  \\ 
      &  & MAE & 1.1205 & 0.9731 & 0.6138 & \textbf{0.2489}   \\ 
    \cmidrule{3-7}
    &  \multirow{2}{*}{Qwen2-7B} & MSE & 2.1457 & 1.6072 & 0.5550 &  \textbf{0.3651}  \\ 
      &  & MAE & 1.1205 & 0.9731/ & 0.4850 & \textbf{0.2838}   \\ 
\bottomrule
\end{tabularx}
\label{tab:general_annotation_evaluation_forecasting_full}
\end{table*}


\subsection{Evaluation in Time Series Imputation Tasks}
\label{appendix:time_series_imputation_more_resuilts}
To demonstrate the effectiveness of \method in improving the performance of various downstream tasks, we further apply the generated general annotations in time series imputation task. Specifically, time series imputation task refers to the process of filling in missing or incomplete data points in a time series dataset, where some values are randomly mask.

\noindent\textbf{Implementation Details}. We implement the multi-modal time series imputation based on TSLib~\cite{wu2023timesnet}. 
We use Informer~\cite{zhou2021informer} as the forecasting model for the time series forecasting task. The model is configured with a dropout rate of $0.1$ and a learning rate of $0.0001$. GPT-4o is set as the LLM backbone. Other settings follow these in Section~\ref{appendix:implementation_details}.

\noindent\textbf{Experimental Results}. The experimental results are shown in Table~\ref{tab:general_annotation_evaluation_imputation}. From the table, we observe that TESSA consistently outperforms baselines in all datasets, demonstrating that TESSA's annotations can significantly benefits various downstream tasks, including forecasting and imputation.

\begin{table*}[t]
\centering
\small
\caption{Imputation results with GPT-4o as the LLM backbone. Informer is the imputation model.}
\begin{tabularx}{0.65\linewidth}{ll|cccc}
\toprule
\textbf{Metric} & \textbf{Domain} & \textbf{NoText} & \textbf{TimeMMD} & \textbf{DirectLLM} & \textbf{TESSA} \\ \midrule
\multirow{3}{*}{MSE} & \multirow{1}{*}{Environment}  & 0.9718
& 0.9657 & 0.9453 & \textbf{0.5698} \\
& \multirow{1}{*}{Energy} & 0.9109 & 0.9081 & 0.9018 & \textbf{0.8690} \\
& \multirow{1}{*}{Social Good}  & 1.4971
& 0.9784 & 0.6873 & \textbf{0.5492}\\
\midrule
\multirow{3}{*}{MAE} & \multirow{1}{*}{Environment}  &  0.6872& 0.6867 & 0.6973 & \textbf{0.5438}  \\
& \multirow{1}{*}{Energy}   & 0.8216 & 0.8176 & 0.8111 & \textbf{0.8075} \\
& \multirow{1}{*}{Social Good}   & 0.8371 & 0.7806 & 0.6036 & \textbf{0.5116}\\
\bottomrule
\end{tabularx}
\label{tab:general_annotation_evaluation_imputation}
\end{table*}

\section{Additional Details of General Annotation Evaluation in Synthetic Datasets}

\subsection{Implementation Details}
\label{appendix:general_annotation_synthetic_implementation_details}
\nop{To use an LLM-as-a-judger to evaluate the effectiveness of \method in generating general annotations for synthetic time series, GPT-4o is set as the backbone of judgers.  Two metrics, \textit{Clarity}, \textit{Comprehensiveness} are applied to evaluate the quality of annotations:
\begin{itemize}[leftmargin=*]
    \item Clarity: it assesses the clarity and readability
    \item Comprehensiveness: it checks whether the annotation cover more important patterns.
\end{itemize}}
To evaluate the effectiveness of \method in generating general annotations for synthetic time series using an LLM-as-a-judger approach, we set GPT-4o as the backbone of the judger. Two metrics, \textit{Clarity} and \textit{Comprehensiveness}, are used to assess the quality of the annotations:
\begin{itemize}[leftmargin=*]
    \item \textbf{Clarity}: Evaluates the clarity and readability of the annotations.
    \item \textbf{Comprehensiveness}: Assesses whether the annotations cover the most important patterns.
\end{itemize}


\section{Additional Results of Domain-specific Annotation Evaluation}

\subsection{Evaluation Metrics}
\label{appendix:domain_specific_annotation_evaluation_metric}
\nop{We use the following three metrics to evaluate the quality of domain-specific annotations:
\begin{itemize}[leftmargin=*]
    \item \textbf{Clarity}: it assesses the clarity and readability
    \item \textbf{Comprehensiveness}: it checks whether the annotation cover more important patterns.
    \item \textbf{Domain-relevance}: it evaluate whether the annotation correctly applies domain-specific knowledge
\end{itemize}}

We use the following three metrics to evaluate the quality of domain-specific annotations:
\begin{itemize}[leftmargin=*]
    \item \textbf{Clarity}: Assesses the clarity and readability of the annotations.
    \item \textbf{Comprehensiveness}: Checks whether the annotations cover the most important patterns.
    \item \textbf{Domain-Relevance}: Evaluates whether the annotations correctly apply domain-specific knowledge.
\end{itemize}


\subsection{Additional Results on Other LLM Backbones}
\label{appendix:domain_specific_evaluation_more_results}
\nop{We report the evaluation results of domain-specific annotations on the Energy and Social Good datasets in Table~\ref{tab:domain_specific_annotation_evaluation_appendix_energy} and Table~\ref{tab:domain_specific_annotation_evaluation_appendix_social}, respectively. Similar observations are found in Section~\ref{sec:domain_specific_annotation_evaluation}, which demonstrate the effectiveness of \method in generating high-quality domain-specific annotations.}

We report the evaluation results of the domain-specific annotations on the Energy and Social Good datasets in Table~\ref{tab:domain_specific_annotation_evaluation_appendix_energy} and Table~\ref{tab:domain_specific_annotation_evaluation_appendix_social}, respectively. Similar observations are made in Section~\ref{sec:domain_specific_annotation_evaluation}, further demonstrating the effectiveness of \method in generating high-quality domain-specific annotations.

\begin{table}[ht]
\centering
\small
\caption{Domain-specific annotation results on the Energy dataset with GPT-4o as the LLM backbone.}

\begin{tabularx}{0.9\linewidth}{llcc}
\toprule
\textbf{Metric} & \textbf{Method} & \textbf{Mean} & \textbf{P(T>D) (\%)} \\ \midrule
\multirow{2}{*}{Clarity}  & TESSA & \textbf{4.79} & \multirow{2}{*}{\textbf{99.35}}
\\
  & DirectLLM & 3.48 & \\ 
\midrule
\multirow{2}{*}{Compre.}  & TESSA & \textbf{4.57} & \multirow{2}{*}{\textbf{98.01}}
\\
  & DirectLLM & 3.10 & \\ 
  \midrule
\multirow{2}{*}{Dom. Rel.}  & TESSA & \textbf{4.25} & \multirow{2}{*}{\textbf{95.24}}
\\
  & DirectLLM & 3.01 & \\ 
  \midrule
\multirow{2}{*}{Overall}  & TESSA & \textbf{4.57 }& \multirow{2}{*}{\textbf{98.31}}
\\
  & DirectLLM & 3.35 & \\ 

\bottomrule
\end{tabularx}
\label{tab:domain_specific_annotation_evaluation_appendix_energy}
\end{table}

\begin{table}[ht]
\centering
\small
\caption{Domain-specific annotation results on the Social Good dataset with GPT-4o as the LLM backbone.}

\begin{tabularx}{0.9\linewidth}{llcc}
\toprule
\textbf{Metric} & \textbf{Method} & \textbf{Mean} & \textbf{P(T>D) (\%)} \\ \midrule
\multirow{2}{*}{Clarity}  & TESSA & \textbf{4.68} & \multirow{2}{*}{\textbf{99.61}}
\\
  & DirectLLM & 3.28 & \\ 
\midrule
\multirow{2}{*}{Compre.}  & TESSA & \textbf{4.49} & \multirow{2}{*}{\textbf{97.54}}
\\
  & DirectLLM & 3.26 & \\ 
  \midrule
\multirow{2}{*}{Dom. Rel.}  & TESSA & \textbf{4.45} & \multirow{2}{*}{\textbf{95.34}}
\\
  & DirectLLM & 3.33 & \\ 
  \midrule
\multirow{2}{*}{Overall}  & TESSA & \textbf{4.48 }& \multirow{2}{*}{\textbf{97.16}}
\\
  & DirectLLM & 3.29 & \\ 

\bottomrule
\end{tabularx}
\label{tab:domain_specific_annotation_evaluation_appendix_social}
\end{table}

\section{Experimental results of Human in Loop}
In this section, we demonstrate that the annotations generated by \method can assist humans in analyzing time-series data. To evaluate this, we selected $60$ time-series samples from three domains—Environment, Energy, and Social Good—with $20$ datasets per domain.
We compared annotations generated by \method and DirectLLM, asking $20$ PhD stduents, researchers, and professors as the participants to assess which annotations were more informative and useful.  
The results of this human-in-the-loop evaluation, summarized in Table~\ref{tab:results_human_in_loop}, reveal that \method consistently outperforms DirectLLM across all three domains. Participant assessments indicate that $88.3\%$ of \method's general annotations and $93.3\%$ of its domain-specific annotations are more informative compared to DirectLLM’s outputs. This substantiates \method’s capacity to produce semantically meaningful annotations that enhance human interpretability during time-series analysis workflows.


\begin{table}[ht]
\centering
\small
\caption{Comparison results of general and domain-specific annotations. GPT-4o is the LLM backbone.}
\begin{tabularx}{0.7\linewidth}{llc}
\toprule
\textbf{Domain} & \textbf{Method} & \textbf{P(T>D) (\%)} \\ \midrule
\multirow{2}{*}{Environment}  & General & {83.3}
\\
  & Specific & 91.7 \\ 
\midrule
\multirow{2}{*}{Energy}  & General & {88.3}
\\
  & Specific & 93.3 \\ 
  \midrule
\multirow{2}{*}{Social Good}  & General & {85.8} 
\\
  & Specific & 92.5 \\ 
\bottomrule
\end{tabularx}
\label{tab:results_human_in_loop}
\end{table}

\section{Additional Results on Multivariate Time Series}
\label{appendix:multivariate_ts_results}
In this section, we aim to demonstrate the effectiveness of \method in generalizing to multivariate time series. Specifically, We use Stock and Health datasets as the source domains, Climate and Economy datasets as the target domains. 

\noindent\textbf{Domain-specific Annotations Evaluation}. Similar to Section~\ref{sec:domain_specific_annotation_evaluation}, we adopt a LLM-as-a-Judger strategy to evaluate the performance of \method and DirectLLM in generating domain-specific annotations. Other settings follow these in Section~\ref{sec:domain_specific_annotation_evaluation}.
We present the comparison results on the two datasets in Table~\ref{tab:domain_specific_annotation_evaluation_appendix_climate_economy}. From the table, we observe that similar to these of univariate time series in Section~\ref{sec:domain_specific_annotation_evaluation}, \method significantly outperforms DirectLLM across all metrics in both two multivariate time series datasets, aciving an overall score of $4.51$ and $4.55$ compared to DirectLLM's $3.38$ and $3.42$ in Climate and Economy dataset, respectively. This demonstrates the effectiveness of \method in generalizing to multivariate time series. 
Additional case studies of TESSA applied to multivariate time series are presented in Appendix~\ref{appendix:case_study}.

\begin{table}[ht]
\centering
\small
\tabcolsep 3.5pt
\renewcommand\arraystretch{1.0}
\caption{Comparison results of domain-specific annotations for multivariate time series in the Climate and Economy datasets using GPT-4o as the LLM backbone.}

\begin{tabular}{lllcc}
\toprule
\textbf{Domains} & \textbf{Metric} & \textbf{Method} & \textbf{Mean} & \textbf{P(T>D) (\%)} \\ \midrule
\multirow{8}{*}{Climate} & \multirow{2}{*}{Clarity}  & TESSA & \textbf{4.58} & \multirow{2}{*}{\textbf{99.17}}
\\
 & & DirectLLM & 3.35 & \\ 
\cmidrule{2-5}
& \multirow{2}{*}{Compre.}  & TESSA & \textbf{4.37} & \multirow{2}{*}{\textbf{96.43}}
\\
&  & DirectLLM &  3.38& \\ 
\cmidrule{2-5}
& \multirow{2}{*}{Dom. Rel.}  & TESSA & \textbf{4.57} & \multirow{2}{*}{\textbf{96.20}}
\\
 & & DirectLLM &  3.40 & \\ 
\cmidrule{2-5}
& \multirow{2}{*}{Overall}  & TESSA & \textbf{4.51}& \multirow{2}{*}{\textbf{97.26}}
\\
 & & DirectLLM & 3.38 & \\ 
\midrule
\multirow{8}{*}{Economy} & \multirow{2}{*}{Clarity}  & TESSA & \textbf{4.46} & \multirow{2}{*}{\textbf{97.14}}
\\
 & & DirectLLM & 3.48 & \\ 
\cmidrule{2-5}
& \multirow{2}{*}{Compre.}  & TESSA & \textbf{4.57} & \multirow{2}{*}{\textbf{96.81}}
\\
&  & DirectLLM &  3.41 & \\ 
\cmidrule{2-5}
& \multirow{2}{*}{Dom. Rel.}  & TESSA & \textbf{4.61} & \multirow{2}{*}{\textbf{96.70}}
\\
 & & DirectLLM & 3.37 & \\ 
\cmidrule{2-5}
& \multirow{2}{*}{Overall}  & TESSA & \textbf{4.55}& \multirow{2}{*}{\textbf{96.88}}
\\
 & & DirectLLM & 3.42 & \\ 
\bottomrule
\end{tabular}
\label{tab:domain_specific_annotation_evaluation_appendix_climate_economy}
\end{table}


\section{Additional Details of Ablation Studies}
\label{appendix:ablation_studies}
\subsection{Evaluation Metric}
\label{appendix:ablation_study_evaluation_metric}
To evaluate the effectiveness of the adaptive feature selection, we use an LLM-as-a-judger to evaluate the general annotations generated by \method and its variant \method/F. 


\subsection{Qualitative Examples}
\label{appendix:ablation_study_qualitative_examples}
We present a qualitative example of extracting text-wise features using \method and \method/D, shown in Table~\ref{tab:ablation_study_extracted_features}. From the table, we observe that \method/D captures irrelevant features, such as \textit{higher prices over time} and \textit{fun}, which are unrelated to time series analysis. This supports our claim that domain-specific terminology can hinder the accurate extraction of time-series-relevant features.


We also provide another qualitative example in Table~\ref{tab:ablation_study_impact_of_feature_selection} to demonstrate the effectiveness of adaptive feature selection. The annotations generated by \method/F tend to include numerous features without proper analysis. This illustrates that including too many features can reduce the clarity of the annotations, further emphasizing the importance of adaptive feature selection in improving annotation quality.

\begin{table}[t]
\small
\centering
\caption{Ablation studies of the impact of domain decontextualization. \colorbox{lightred}{Red} denotes the irrelevant features.}
\begin{tabular}{p{\linewidth}} \toprule
\textbf{TESSA's extracted text-wise features}: 

support level, resistance level, volume correlation, breakthrough, trend reversal, relative strength index, negative signal, positive signal, channel boundaries

\\ \midrule
\textbf{TESSA/D's extracted text-wise features}:

\colorbox{lightred}{higher prices over time}, autocorrelation, \colorbox{lightred}{price increase}, trend channel, stationary, \colorbox{lightred}{fun}, \colorbox{lightred}{lower prices}, outliers, breakdown, rising trend

\\ \bottomrule
\end{tabular}
\label{tab:ablation_study_extracted_features}
\end{table}

\begin{table}[ht]
\centering
\small
\caption{Case study: Evaluation results of domain-specific annotation of time series data in Fig.~\ref{fig:case_study_ts_data} from the Environment dataset. GPT-4o is the LLM backbone.}
\begin{tabularx}{0.7\linewidth}{llc}
\toprule
\textbf{Metric} & \textbf{Method} & \textbf{Score} \\ \midrule
\multirow{2}{*}{Clarity}  & TESSA & \textbf{5.0}
\\
  & DirectLLM & 3.0 \\ 
\midrule
\multirow{2}{*}{Compre.}  & TESSA & \textbf{3.0}
\\
  & DirectLLM & 3.0 \\ 
  \midrule
\multirow{2}{*}{Dom. Rel.}  & TESSA & \textbf{5.0} 
\\
  & DirectLLM & 3.0 \\ 
  \midrule
\multirow{2}{*}{Overall}  & TESSA & \textbf{4.3 }
\\
  & DirectLLM & 3.0 \\
\bottomrule
\end{tabularx}
\label{tab:case_study_domain_specific_evaluation_main}
\end{table}

\begin{table*}[t]
\small
\centering
\caption{Ablation studies of the impact of adaptive feature selection. Two compared methods are used for generating general annotation ploted in Fig.~\ref{fig:case_study_ts_data}(a). \colorbox{lightpurple}{Purple} highlights basic patterns, \colorbox{lightgreen}{green} highlights general patterns.}
\begin{tabular}{p{\linewidth}} \toprule
\textbf{TESSA's generated general annotation}: 

The time series data tracks a single variable over time, showing a gradual \colorbox{lightpurple}{upward trend} from around 6.96 to over 10.4. There are clear \colorbox{lightgreen}{seasonal} patterns, with regular cycles of peaks and troughs, hinting at \colorbox{lightgreen}{periodic} influences like seasonal demand. The data exhibits strong \colorbox{lightgreen}{support and resistance levels}, where it hesitates at specific values before continuing its rise, demonstrating \colorbox{lightgreen}{resilience}. There is a consistent \colorbox{lightpurple}{upward indicator trend}, suggesting stable or improving underlying factors driving this increase. Overall, the data indicates a \colorbox{lightgreen}{positive growth trajectory}, with predictable cyclical fluctuations due to natural or recurring factors.

\\ \midrule
\textbf{TESSA/F's generated general annotations}:

The provided time series data shows a \colorbox{lightpurple}{clear upward trend}, indicated by the gradual increase in \colorbox{lightpurple}{trend} values over time. Initially, \colorbox{lightgreen}{fluctuations} are minor, but as the series progresses, we observe larger increases, especially towards the end of the sequence, suggesting a \colorbox{lightgreen}{strong positive momentum}. The \colorbox{lightgreen}{occasional deviations} from the trend, as shown by the \colorbox{lightgreen}{residue}, hint at \colorbox{lightgreen}{short-term fluctuations} or potential \colorbox{lightgreen}{anomalies}. Seasonality seems to have minimal impact with slight \colorbox{lightgreen}{cyclical} patterns, possibly reflecting \colorbox{lightgreen}{periodic} but non-dominant effects. Overall, if this time series pertains to a scenario where higher values are desirable, such as sales or \colorbox{lightgreen}{economic indicators}, this upward trend and increased values could be considered as positive signals or a \colorbox{lightgreen}{breakthrough}.

\\ \bottomrule
\end{tabular}
\label{tab:ablation_study_impact_of_feature_selection}
\end{table*}

\subsection{Additional Ablation Studies on the Contributions of \method's Components}
\label{appendix:more_ablation_studies_tessa_components}
In Section~\ref{sec:in_depth_TESSA}, we conduct ablation studies to assess the importance of domain decontextualization and adaptive feature selection in \method. To further understand the contributions of \method's components, we conduct additional ablation studies on domain-specific term extractor and annotation reviewers, respectively. 

\noindent\textbf{Domain-specific Term Extractor}. To understand the ability of LLMs to process domain-specific jargon to generate useful features, we implement a variant \method/S, which removes the domain-specific term extractor in \method and directly generate domains-specific annotations from general annotations. 
We provide the comparison results in Stock and SocialGood datasets in Tables~\ref{tab:ablation_study_jargon} and~\ref{tab:ablation_study_jargon1}, which demonstrate that \method is able to capture more jargon-rich in information, such as resistance levels in the Stock dataset and reactive bounds in the SocialGood dataset. In contrast, \method/S merely converts general features into basic domain-specific features. This highlights that LLMs, when used as domain-specific term extractors, can effectively generate valuable jargon that enhances the generation of domain-specific annotations.

\noindent\textbf{Annotation Reviewer}. To evaluate the impact of the annotation reviewer component, we introduce a variant \method/A that excludes this module. We assess the quality of domain-specific annotations generated by \method and \method/A using an LLM-as-a-judger framework, focusing on two criteria: (1) \textit{Clarity} (readability and coherence of annotations) and (2) \textit{Domain-Relevance} (alignment with domain-specific context). An overall score, calculated as the average of these metrics, provides a holistic performance measure.
As shown in Table~\ref{tab:appendix_ablation_studies_annotation_reviewer}, \method achieves statistically significant improvements over \method/A across all metrics. This quantitative comparison demonstrates that the annotation reviewer critically enhances both the clarity and contextual relevance of generated annotations, ensuring they better capture domain-specific nuances.


\begin{table}[t]
\small
\centering
\caption{Ablation studies of the impact of the domain-specific term extractor in Stock dataset. \colorbox{lightpurple}{Purple} highlights basic patterns, \colorbox{lightgreen}{green} highlights general patterns, \colorbox{yellow}{yellow} highlights domain-specific patterns and \colorbox{lightblue}{blue} highlights correlations between variables in multivariate time series data.}
\begin{tabular}{p{\linewidth}} \toprule
\textbf{TESSA}: 

Compass Digital Acquisition Unit's stock price shows a notable pattern of \colorbox{lightgreen}{rising and falling periodically}, indicating \colorbox{lightgreen}{seasonality} with stable \colorbox{yellow}{long-term trend}s interrupted by short-term \colorbox{lightpurple}{fluctuations}. There are key \colorbox{yellow}{resistance levels} around intervals 134,270, and 403, where prices peak before dipping. The stock volume demonstrates significant spikes at specific points, suggesting irregular activity, particularly around values 7,000 and 80,500, which may indicate \colorbox{yellow}{volume bursts} or unusual market events. The relative strength index (RSI) also reveals a \colorbox{lightgreen}{recurring} pattern, gradually \colorbox{lightpurple}{trending upward} before a sharp decline, reflecting a cycle of growth and subsequent drop. Overall, the mild \colorbox{lightblue}{positive correlation between stock price and RSI} indicates that periodic changes in price are somewhat echoed in RSI patterns, potentially offering predictive insights for future stock movements. 
\\ \midrule
\textbf{TEESA/S}:

Compass Digital Acquisition Unit's stock price exhibits a distinct \colorbox{lightgreen}{cyclical pattern}, suggesting \colorbox{lightgreen}{seasonality}, with occasional \colorbox{lightpurple}{fluctuations}. The relative strength index (RSI) shows a \colorbox{lightgreen}{repeating} trend, gradually \colorbox{lightpurple}{rising} before experiencing a sharp drop, indicating a pattern of growth followed by a decline. The mild \colorbox{lightblue}{positive relationship between stock price and RSI} suggests that changes in price tend to be mirrored in the RSI, offering potential clues for predicting future price movements.

\\ \bottomrule
\end{tabular}
\label{tab:ablation_study_jargon}
\end{table}

\begin{table}[t]
\small
\centering
\caption{Ablation studies of the impact of the domain-specific term extractor in SocialGood dataset. \colorbox{lightpurple}{Purple} highlights basic patterns, \colorbox{lightgreen}{green} highlights general patterns and \colorbox{yellow}{yellow} highlights domain-specific patterns.}
\begin{tabular}{p{\linewidth}} \toprule
\textbf{TESSA}: 
The unemployment rate displays a \colorbox{lightpurple}{clear upward} trend over time, starting from the mid-5s and progressing to the mid-7s toward the end of the series, signaling \colorbox{lightgreen}{persistent growth} in unemployment. This movement is influenced by \colorbox{lightgreen}{volatility periods}, suggesting that economic cycle phases are periodically impacting employment levels. The observed \colorbox{yellow}{resistance thresholds} around the mid-5s and mid-6s mark critical \colorbox{yellow}{threshold levels}, where unemployment temporarily stabilizes before continuing its upward momentum. Analyzing \colorbox{lightgreen}{lag features}, past values like 5.5 and 6.0 serve as baselines, helping to understand how unemployment has evolved over time. This steady upward movement, punctuated by cyclical variations, highlights the need for strategies that address both \colorbox{yellow}{economic momentum indicators} and short-term \colorbox{yellow}{reactive bounds}, ensuring a robust response to both long-term trends and periodic disruptions in the labor market.
\\\midrule
\textbf{TEESA/S}:
The unemployment rate data reveals a consistent \colorbox{lightpurple}{upward trend} over time, starting from 5.18 and rising to approximately 7.24. This suggests an \colorbox{lightpurple}{increasing trend} in unemployment, which might pose socio-economic challenges if it persists. The data also exhibits \colorbox{lightgreen}{seasonal fluctuations}, indicating periods of elevated unemployment that could align with specific economic or legislative cycles. These recurring ups and downs highlight how external factors might \colorbox{lightgreen}{periodically} impact the job market. Recognizing these \colorbox{lightgreen}{cyclical variations} can help policymakers craft effective strategies to mitigate the potential socio-economic risks of rising unemployment in the future.
\\ \bottomrule
\end{tabular}
\label{tab:ablation_study_jargon1}
\end{table}

\begin{table}[ht]
\centering
\small
\caption{Ablation studies for the annotation reviewer in the Social Good dataset with GPT-4o as the LLM backbone.}
\begin{tabularx}{0.95\linewidth}{llcc}
\toprule
\textbf{Metric} & \textbf{Method} & \textbf{P(TESSA>TESSA/A) (\%)} \\ \midrule
\multirow{2}{*}{Clarity}  & TESSA & \multirow{2}{*}{\textbf{95.4}}
\\
  & \method/A & \\ 
\midrule
\multirow{2}{*}{Dom. Rel.}  & TESSA  & \multirow{2}{*}{\textbf{87.6}}
\\
  & \method/A  & \\ 
  \midrule
\multirow{2}{*}{Overall}  & TESSA & \multirow{2}{*}{\textbf{91.6}}
\\
  & \method/A & \\ 

\bottomrule
\end{tabularx}
\label{tab:appendix_ablation_studies_annotation_reviewer}
\end{table}



\begin{table}[t]
\centering
\small
\begin{tabular}{l@{}c@{}c@{}c@{}c@{}}
\toprule
\textbf{ Source } & \textbf{ Type } & {\begin{tabular}[c]{@{}c@{}}\textbf{ UTF-8}\\\textbf{bytes}\\\textit{(GB)}\end{tabular}} & {\begin{tabular}[c]{@{}c@{}}\textbf{ Docs }\\\textit{(millions) }\end{tabular}} & {\begin{tabular}[c]{@{}c@{}}\textbf{Tokens}\\\textit{(billions) }\end{tabular}}  \\
\midrule
Common Crawl & web pages & 9,812 & 3,734 & 2,180  \\
GitHub & code & 1,043 & 210 & 342  \\
Reddit & social media & 339 & 377 & 80  \\
Semantic Scholar & papers & 268 & 38.8 & 57  \\
Project Gutenberg & books & 20.4 & 0.056 & 5.2 \\
Wikipedia & encyclopedic & 16.2 & 6.2 & 3.7 \\
\midrule
\multicolumn{2}{c}{\textbf{Total}} & \textbf{11,519} & \textbf{4,367} & \textbf{2,668}  \\
\bottomrule
\end{tabular}
\caption{The categority of the training data used to pre-train OLMo-7B~\cite{groeneveld2024olmo}, which is from OLMo's technical report~\cite{groeneveld2024olmo}.}
\label{tab:olmo_training_data_category}
\end{table}

\subsection{Additional Ablation Studies of Data Contamination}
\label{appendix:ablation_studies_fully_open_LLMs}
To investigate whether \method retains its effectiveness when the LLMs used are not exposed to data describing the time series, we conduct an ablation study using fully open LLMs, specifically OLMo-7B~\cite{groeneveld2024olmo}. The categories of the training data for the LLM are listed in Table~\ref{tab:olmo_training_data_category}.
In this study, we carefully select time series with textual annotations that neither appear in the training data of OLMo-7B nor describe the time series themselves. We use the Stock and Health datasets as source domains, and the Energy dataset as the target domain.
We then apply \method and DirectLLM to generate both general and domain-specific annotations for the time series data in the Energy dataset. 

The results are presented in Table~\ref{tab:olmo_ablation_studies}. From the table, we observe that: (1) despite using a fully open LLM that has not been exposed to data describing the time series, \method with OLMo-7B as the LLM backbone is still able to understand and generate informative annotations with richer general and domain-specific patterns, using more natural language. In contrast, DirectLLM only offers a simplistic description of the basic trend of the time series data. This further underscores the effectiveness of \method even in a scenario of strict data non-contamination.

Additionally, we employ an LLM-as-a-Judger strategy to evaluate the domain-specific annotations generated by \method and DirectLLM, as shown in Table~\ref{tab:olmo_ablation_studies}. The evaluation is conducted from three perspectives: Clarity, Comprehensiveness, and Domain-relevance. GPT-4o serves as the LLM judger, and the other settings follow those in Sec.~\ref{sec:domain_specific_annotation_evaluation}. The comparison results are presented in Table~\ref{tab:domain_specific_ablation_studies_data_contamination}. We observe that \method consistently achieves full scores in all three metrics, significantly outperforming DirectLLM. This further demonstrates \method’s effectiveness in generating high-quality domain-specific annotations with stronger clarity, comprehensiveness, and domain-relevant contextual precision.

\begin{table*}[t]
\small
\centering
\caption{Ablation studies on data contamination using OLMo-7B-Instruct. We select a time-series in the Energy dataset, which is shown in Fig.~\ref{fig:ablation_study_stock_olmo}.
\colorbox{lightpurple}{Purple} highlights basic patterns, \colorbox{lightgreen}{green} highlights general patterns, and \colorbox{yellow}{yellow} highlights domain-specific patterns.}
\begin{tabular}{p{0.95\linewidth}} \toprule
\textbf{TESSA's general annotations} 
The time series data shows a \colorbox{lightpurple}{trend of increasing and decreasing} values with some \colorbox{lightgreen}{seasonality}, which can be observed from the values of the intra-variable time series features such as trend, \colorbox{lightgreen}{seasonality}, and \colorbox{lightgreen}{moving average}. The values of the intra-variable moving average are not stable, indicating that the time series data is \colorbox{lightgreen}{volatile} and prone to \colorbox{lightpurple}{fluctuations}. The values of the intra-variable \colorbox{lightgreen}{lag features} are not significant, indicating that there is no strong correlation among the time series data points.

\textbf{DirectLLM's general annotations} 
This time series data represents a process with high \colorbox{lightpurple}{variability} in the values, with an \colorbox{lightpurple}{increasing trend}. The data is separated by '|' tokens, which could represent different stages or conditions of the process. The highest value of 1.108 is observed at the 11th instance, which might indicate a bottleneck or \colorbox{lightpurple}{peak} in the process. The series shows a \colorbox{lightpurple}{decreasing trend} from instances 9-12, which suggests that certain measures or interventions were put in place to address the bottleneck. Overall, the data suggests that the process is subject to \colorbox{lightpurple}{fluctuations}, but can be managed by identifying and addressing the root causes of the bottlenecks.

\\ \midrule

\textbf{TESSA's domain-specific annotations} 
The natural gas price is exhibiting a \colorbox{lightgreen}{volatile} pattern with significant fluctuations, suggesting a lack of stability in the market. This \colorbox{lightgreen}{volatility} hints at \colorbox{yellow}{underlying market dynamics} that are affecting price movements, potentially tied to factors such as \colorbox{yellow}{supply and demand shifts} or \colorbox{yellow}{external economic variables}. A closer examination of the trend channels indicates intermittent \colorbox{lightpurple}{rising and falling trends}, pointing towards a potential cyclical behavior in natural gas prices. Currently, the market does not show strong correlation from lag features, suggesting that recent \colorbox{lightpurple}{price changes} are not strongly influenced by past values. \colorbox{lightgreen}{Support and resistance levels} could play a crucial role, as breaking through these levels may signal a significant \colorbox{yellow}{trend shift} in the gas market's future pricing.

\textbf{DirectLLM's domain-specific annotations} 
Gasonline prices in the Energy domain are \colorbox{lightpurple}{fluctuating}, ranging from $1.04$ to $1.10$ per unit, with occasional spikes above $1.10$, and drops below $1.00$. These prices \colorbox{lightpurple}{fluctuate} consecutively, indicating a \colorbox{yellow}{dynamic market} for Gasonline.

\\ \bottomrule
\end{tabular}
\label{tab:olmo_ablation_studies}
\end{table*}

\begin{figure}[t]
    \small
    \centering
    \includegraphics[width=0.7\linewidth]{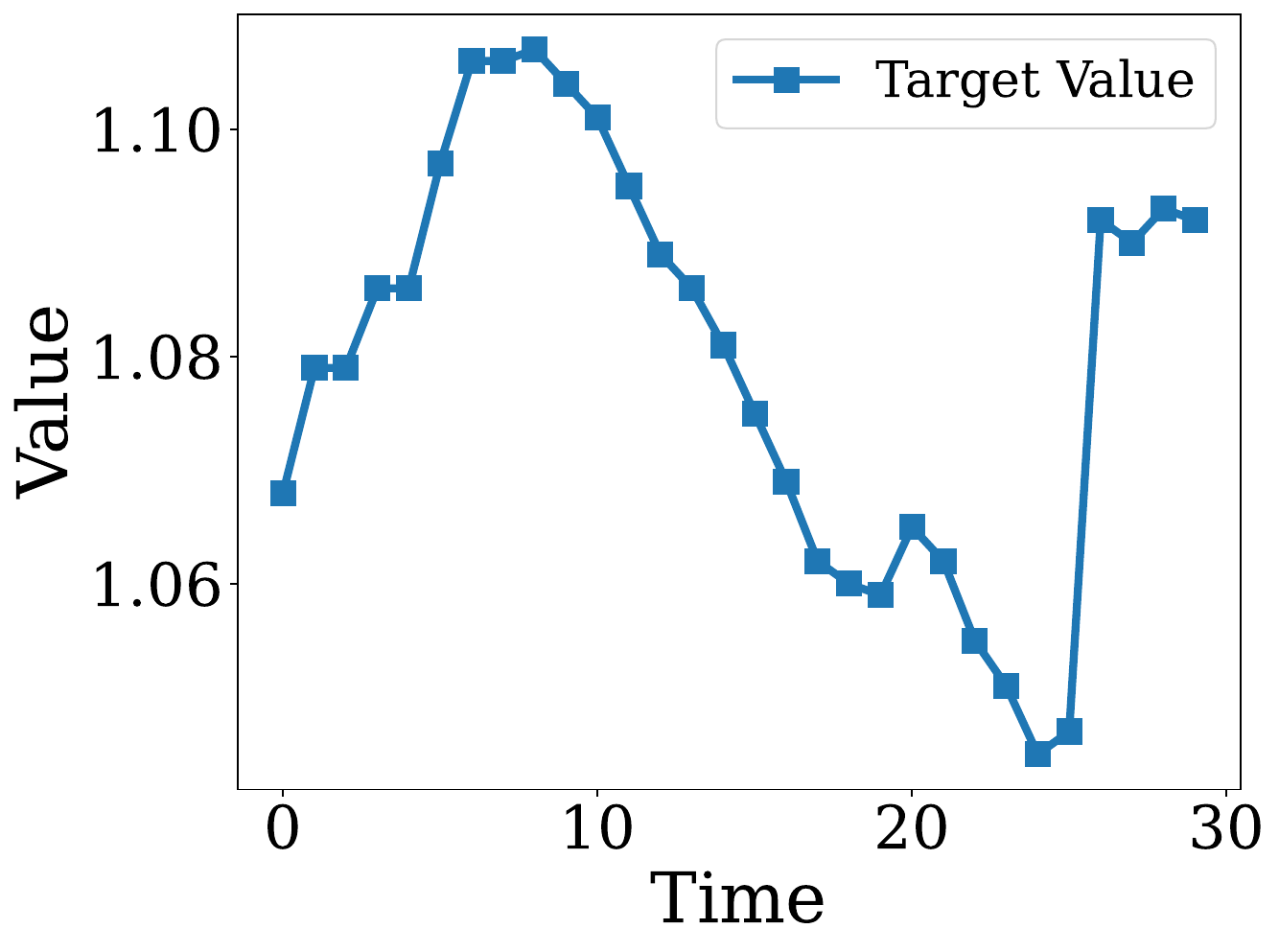}
        \vskip -0.5em
        \caption{Selected time series data from Energy dataset for ablation studies on data contimination. This time series data has 30 data points. The corresponding generated annotations of \method and DirectLLMs are provided in Table~\ref{tab:olmo_ablation_studies}.}
    \vskip -.5em
    \label{fig:ablation_study_stock_olmo}
\end{figure}

\begin{table}[ht]
\centering
\small
\caption{Ablation studies of data contamination. We use an LLM-as-a-Judger to compare a domain-specific annotation in the Energy dataset generated by DirectLLM and \method. OLMo-7B is the LLM-backbone. GPT-4o is used as the LLM judger.}

\begin{tabularx}{0.65\linewidth}{llc}
\toprule
\textbf{Metric} & \textbf{Method} & \textbf{Mean} \\ \midrule
\multirow{2}{*}{Clarity}  & TESSA & \textbf{5} 
\\
  & DirectLLM & 3 \\ 
\midrule
\multirow{2}{*}{Compre.}  & TESSA & \textbf{5} 
\\
  & DirectLLM & 2 \\ 
  \midrule
\multirow{2}{*}{Dom. Rel.}  & TESSA & \textbf{5} 
\\
  & DirectLLM & 3 \\ 
  \midrule
\multirow{2}{*}{Overall}  & TESSA & \textbf{5}
\\
  & DirectLLM & 2.67 \\ 

\bottomrule
\end{tabularx}
\label{tab:domain_specific_ablation_studies_data_contamination}
\end{table}

\section{Additional Details of Case Studies}
\label{appendix:case_study}

\begin{figure}[t]
    \small
    \centering
    \begin{subfigure}{0.49\linewidth}   
        \includegraphics[width=0.98\linewidth]{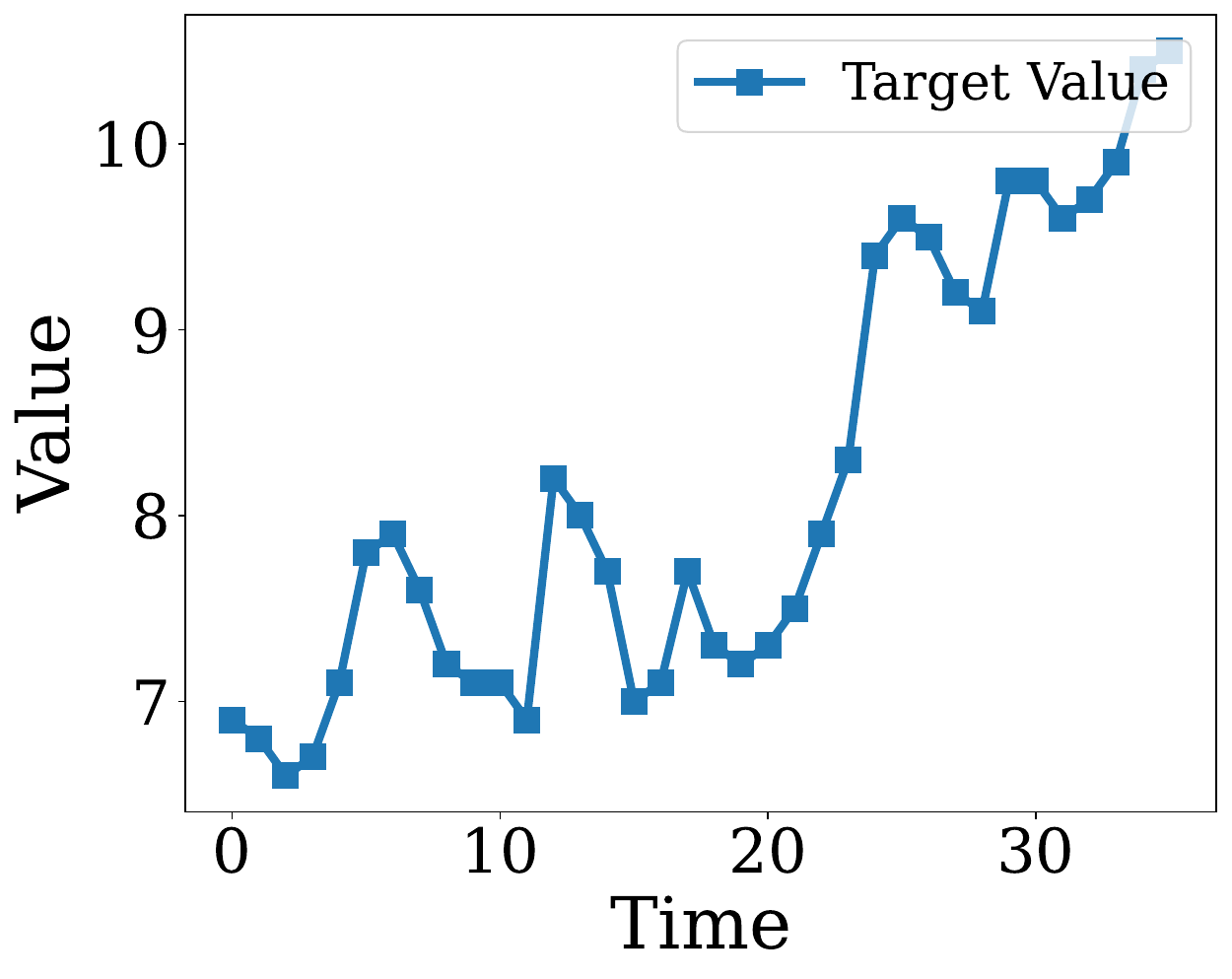}
        \caption{Ablation study}
    \end{subfigure}
    \begin{subfigure}{0.49\linewidth}
        \includegraphics[width=0.98\linewidth]{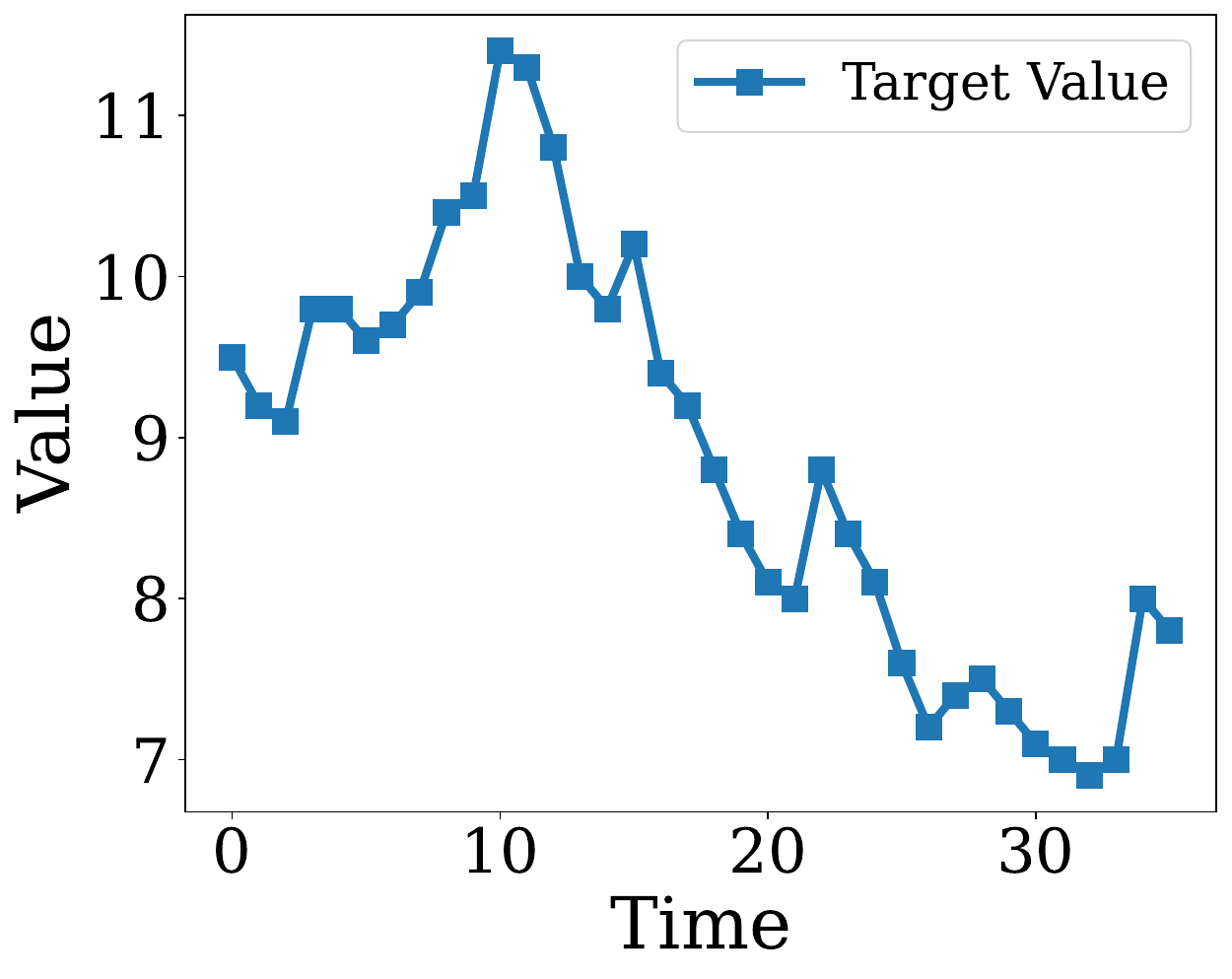}
        \caption{Case study}
    \end{subfigure}
    \caption{Selected time series data from Social Good dataset for ablation studies and case studies. (a) for ablation studies ; and (b) is for case studies.}.
    \label{fig:case_study_ts_data}
\end{figure}

\noindent\textbf{More Details of the Case Studies in Section~\ref{sec:case_study_main}}
\nop{In this section, we provide additional details of case studies. We select a representative time series data from Social Good domain, which is shown in Fig.~\ref{fig:case_study_ts_data}(b). In Table~\ref{tab:case_study_main}, general annotations and domain-specific annotations of both DirectLLM and \method are reported. We further quantitatively evaluate the domain-specific annotations of \method and DirectLLM by following the setting in Section~\ref{sec:domain_specific_annotation_evaluation}. The evaluation results are in Table~\ref{tab:case_study_domain_specific_evaluation_main}.

From the table, we observe that (1) \method's general annotations capture more meaningful patterns, aiding user understanding and downstream tasks, whereas DirectLLM only highlights basic trends; and (2) \method's domain-specific annotations consistently outperform DirectLLM across all metrics, offering clearer, more comprehensive, and contextually relevant insights.}
In this section, we provide additional details of the case studies in Section~\ref{sec:case_study_main}. We select a representative time series from the Social Good domain, shown in Fig.~\ref{fig:case_study_ts_data}(b). In Table~\ref{tab:case_study_main}, both the general and domain-specific annotations generated by DirectLLM and \method are reported. We also quantitatively evaluate the domain-specific annotations of \method and DirectLLM, following the setup outlined in Section~\ref{sec:domain_specific_annotation_evaluation}. The evaluation results are presented in Table~\ref{tab:case_study_domain_specific_evaluation_main}.

From the table, we observe that (1) \method's general annotations capture more meaningful patterns, enhancing user understanding and supporting downstream tasks, while DirectLLM only highlights basic trends; and (2) \method's domain-specific annotations consistently outperform those of DirectLLM across all metrics, providing clearer, more comprehensive, and contextually relevant insights. Specifically, \method's annotations are more fluent, more detailed and provide a richer analysis using domain-specific jargons, like \textit{economic momentum} and \textit{labor market resilience}, while the annotations of DirectLLM only simply analyze the trend of the unemployment rate, providing less insights.

\begin{figure}[t]
    \small
    \centering
    \begin{subfigure}{0.49\linewidth}   
        \includegraphics[width=0.98\linewidth]{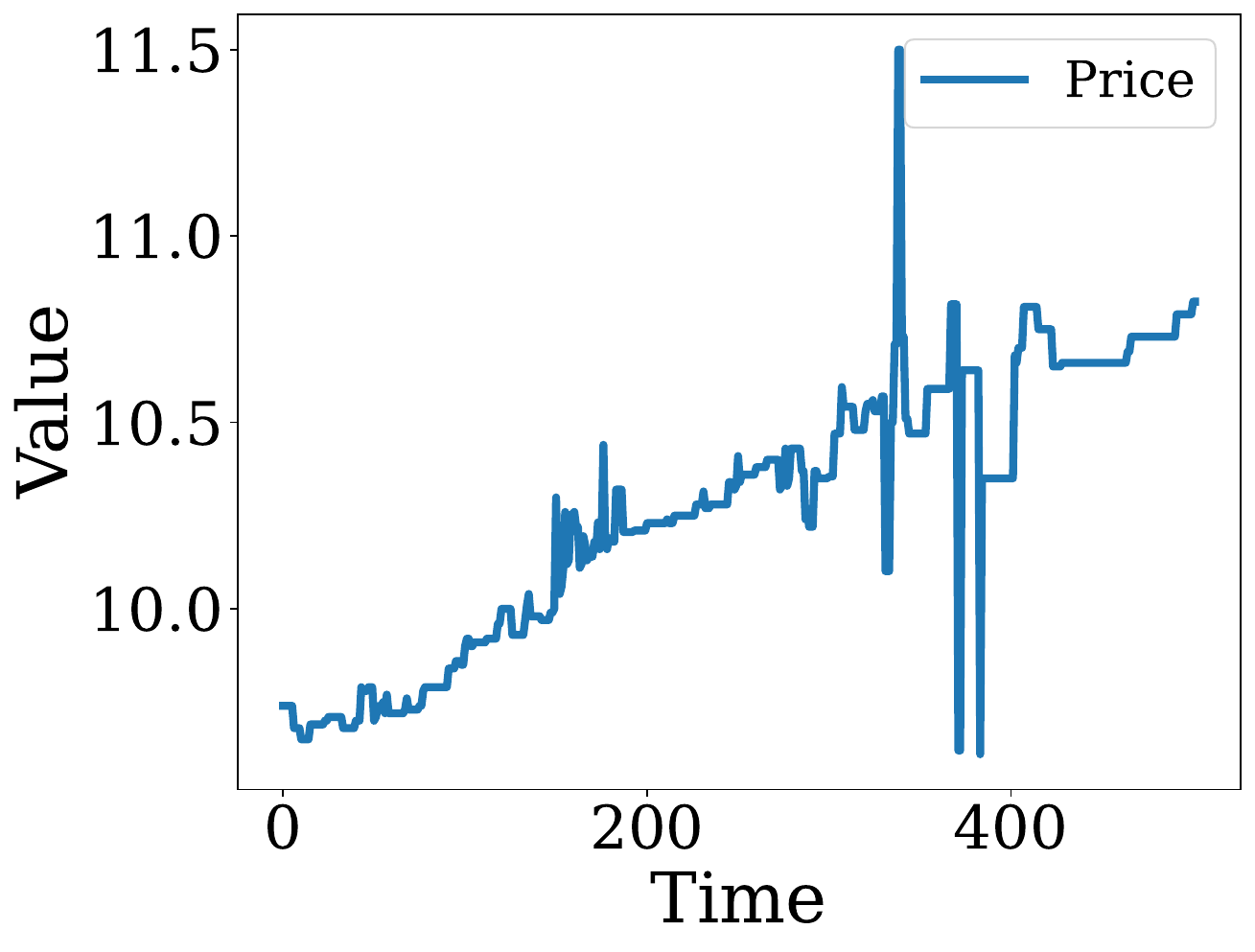}
        \vskip -0.5em
        \caption{Price}
    \end{subfigure}
    \begin{subfigure}{0.49\linewidth}   
        \includegraphics[width=0.98\linewidth]{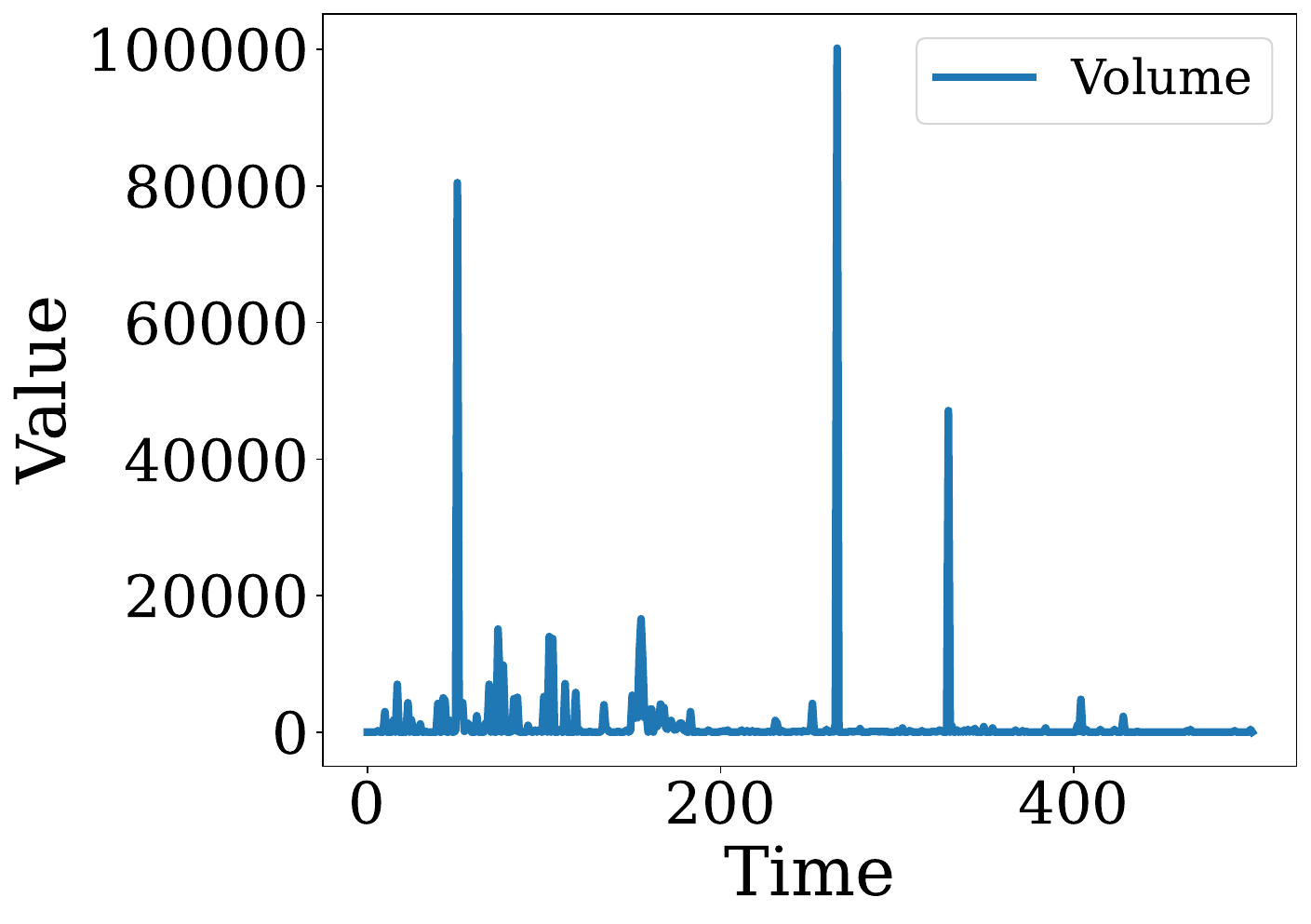}
        \vskip -0.5em
        \caption{Volume}
    \end{subfigure}
    \begin{subfigure}{0.49\linewidth}   
        \includegraphics[width=0.98\linewidth]{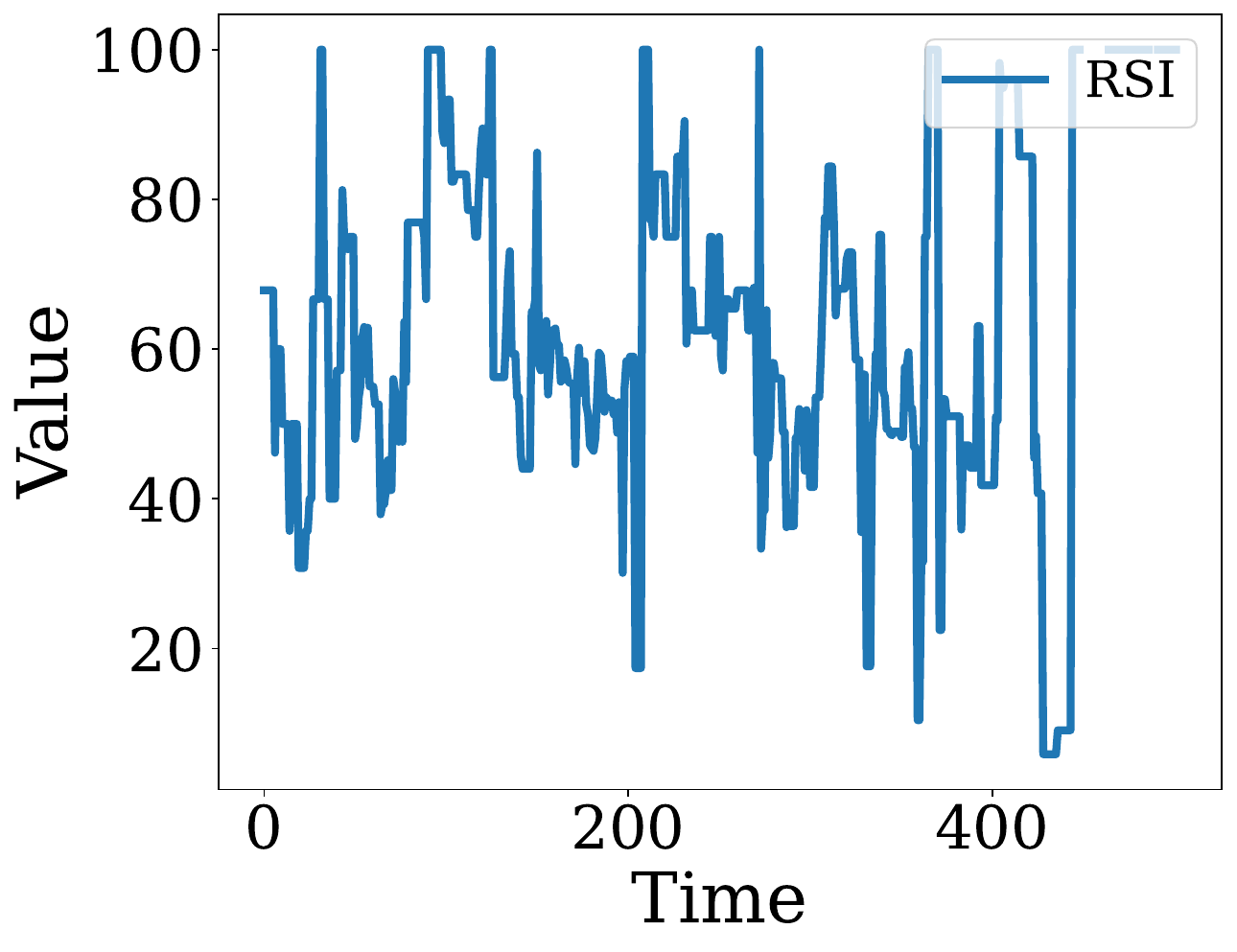}
        \vskip -0.5em
        \caption{RSI}
    \end{subfigure}
    \begin{subfigure}{0.49\linewidth}   
        \includegraphics[width=0.98\linewidth]{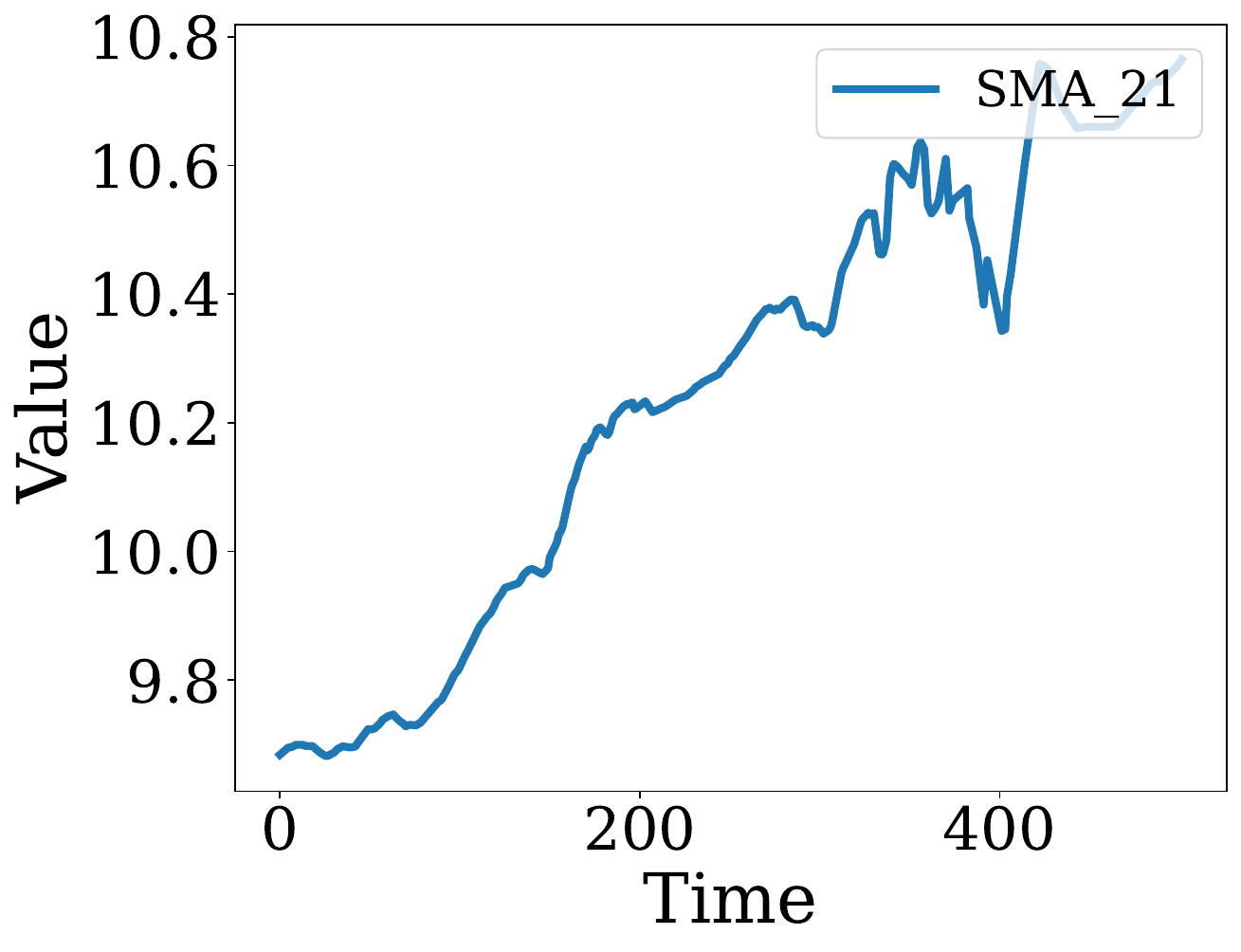}
        \vskip -0.5em
        \caption{SMA}
    \end{subfigure}
    \caption{Case study: a multivariate time series data from the Stock dataset, which has four variables, i.e., price, volumn, RSI and SMA.}.
    \label{fig:case_study_ts_multivariate_data}
\end{figure}

\noindent\textbf{Case Study for Multivariate Time Series}
We then conduct a case study to demonstrate the effectiveness of \method in generating high-quality annotations for multivariate time series data. Specifically, we set the Stock dataset as the target domain. Health and Environment datasets are then applied in the source domains. The example multivariate time series data is shown in Fig.~\ref{fig:case_study_ts_multivariate_data}, where the multivariate time series data has four variables, i.e., price, volume, relative strength index (RSI) and simple moving average (SMA). The generated annotations are shown in Table~\ref{tab:case_study_multivariate}. From the table, we observe that (1) \method's generated annotations are more natural than DirectLLM; (2)
DirectLLM interprets each variable independently by only focusing their trends. However, \method can capture the correlation between variables. This shows \method is able to analyze inter-variable patterns. These further imply the effectiveness of \method in generating high-quality domain-specific annotations for multivariate time series data.

\begin{figure}[t]
    \small
    \centering
    \includegraphics[width=0.98\linewidth]{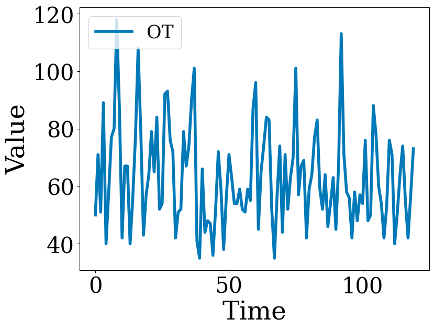}
    \caption{Case study: A selected time series data from the synthetic dataset. The time series data has 120 data points. OT denotes the target variable.}
    \label{fig:case_study_ts_synthetic_data_main}
\end{figure}

\noindent\textbf{Case Study in the Synthetic Dataset}.
We further select an example from the synthetic dataset to conduct similar experiments to generate general annotations.
The selected time series data is in Fig.~\ref{fig:case_study_ts_synthetic_data_main}. The qualitative example of the annotations of this time series data is shown in Table~\ref{tab:case_study_synthetic_more_0}. From the table, we observe a discrepancy in DirectLLM's analysis, as it detects 138 values in the time series data, despite there being only 120 values. This leads to inaccurate annotations. Moreover, DirectLLM captures only the basic trend of the time series, whereas TESSA identifies more significant patterns, such as the \textit{rolling window feature}, \textit{seasonality}, and \textit{resilience}. This demonstrates the effectiveness of \method in providing more comprehensive and accurate annotations. We analyze the reason \method mitigates the hallucination seen in DirectLLM is that it highlights important patterns overlooked by LLMs, such as seasonality. By focusing on these patterns rather than just basic trends, LLMs can analyze and interpret time series data from multiple perspectives, leading to fewer hallucinations in the annotations.

\begin{figure}[t]
    \small
    \centering
    \begin{subfigure}{0.49\linewidth}   
        \includegraphics[width=0.98\linewidth]{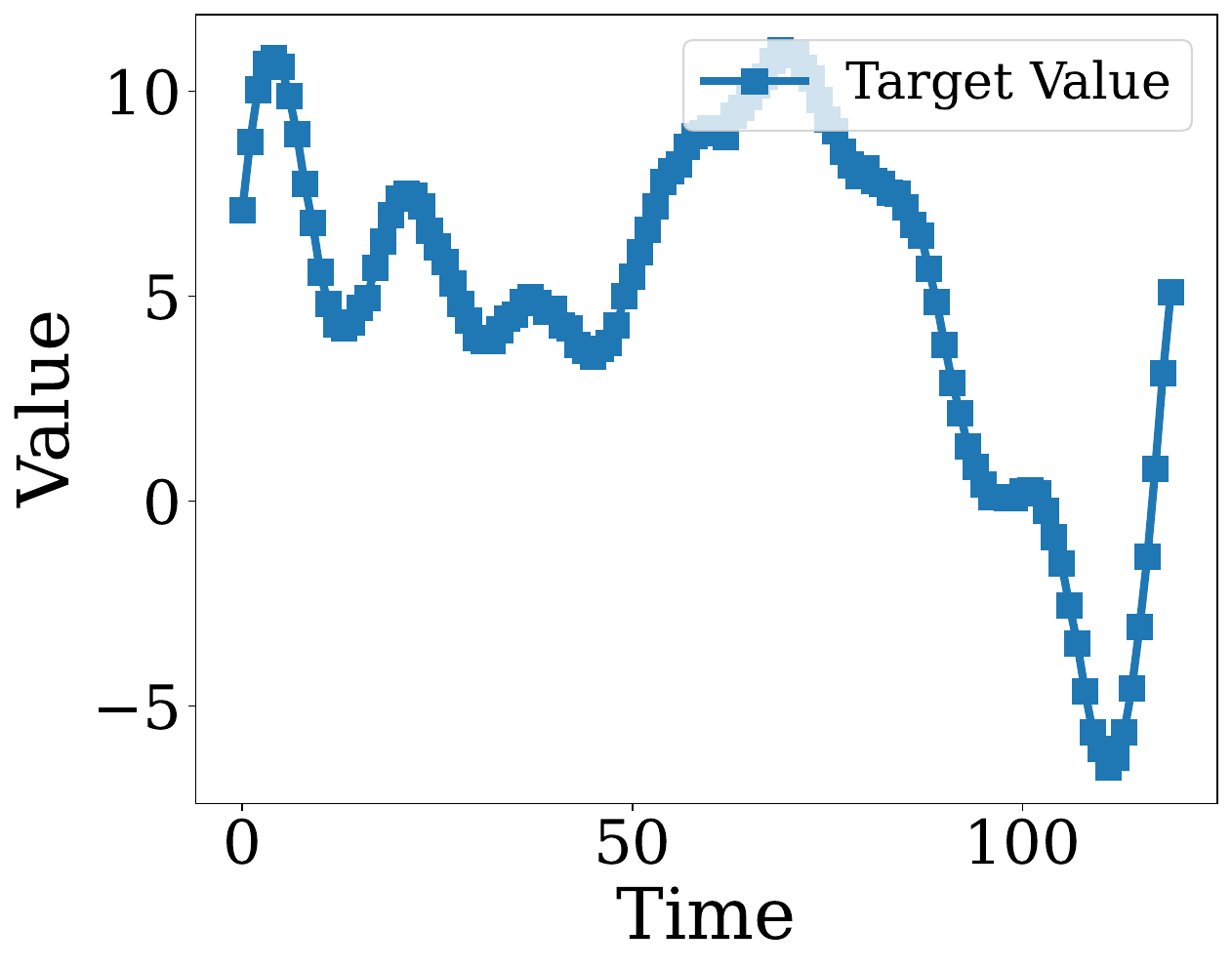}
        \caption{}
    \end{subfigure}
    \begin{subfigure}{0.49\linewidth}   
        \includegraphics[width=0.98\linewidth]{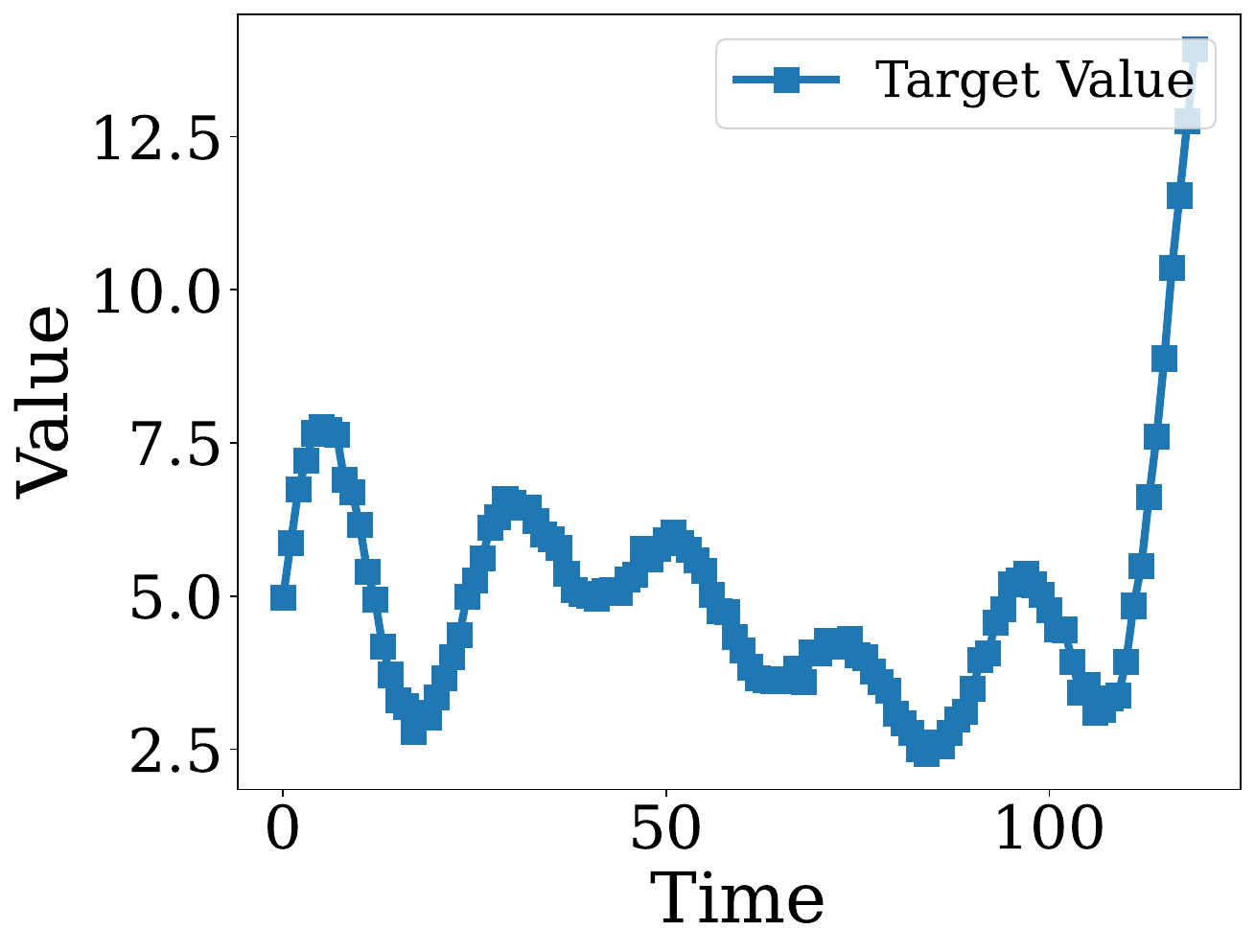}
        \caption{}
    \end{subfigure}
    \caption{More selected time series data from the synthetic dataset. The time series data has 120 data points.}
    \label{fig:case_study_ts_synthetic_data}
\end{figure}

\begin{figure}[t]
    \small
    \centering
    \begin{subfigure}{0.49\linewidth}   
        \includegraphics[width=0.98\linewidth]{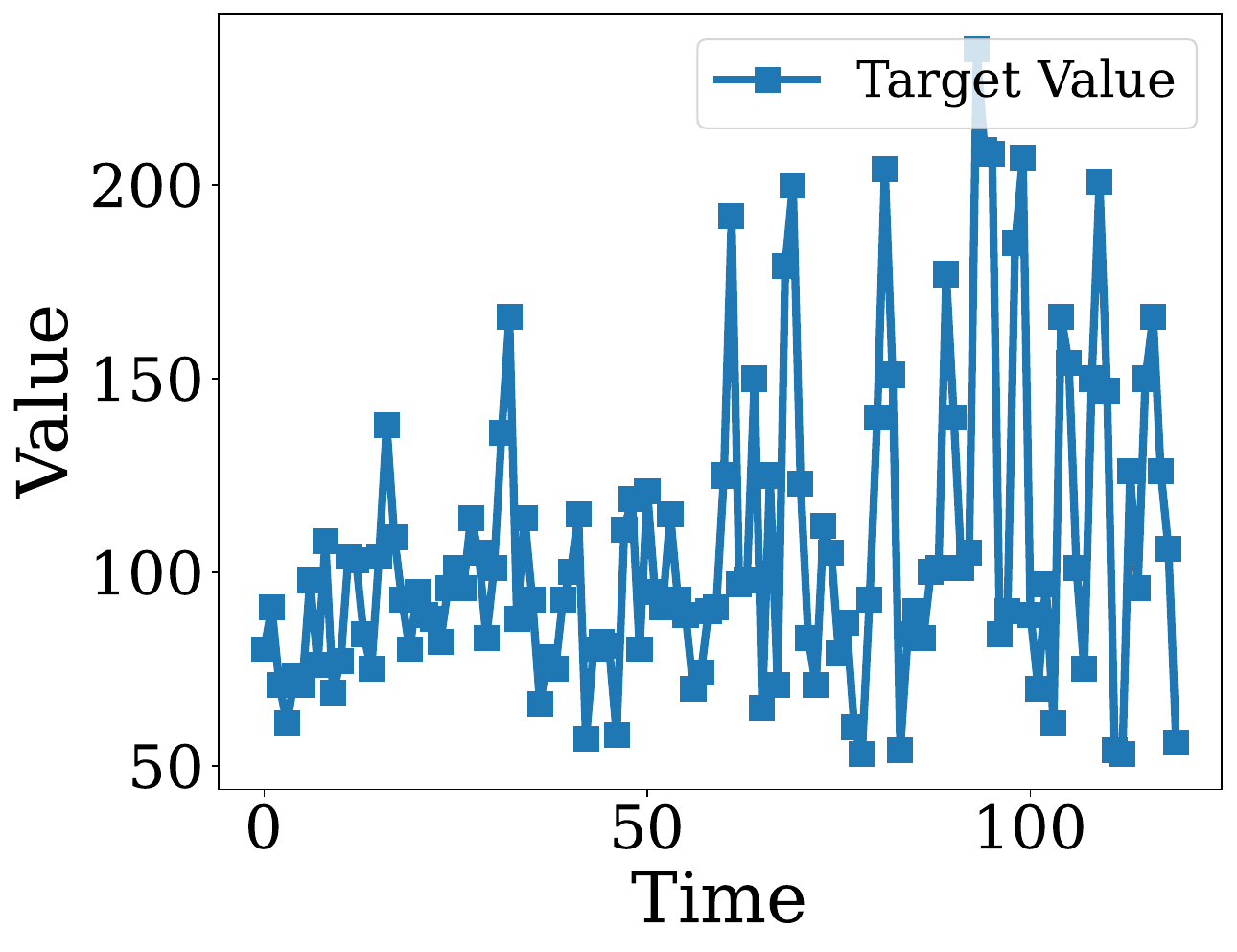}
        \caption{}
    \end{subfigure}
    \begin{subfigure}{0.49\linewidth}   
        \includegraphics[width=0.98\linewidth]{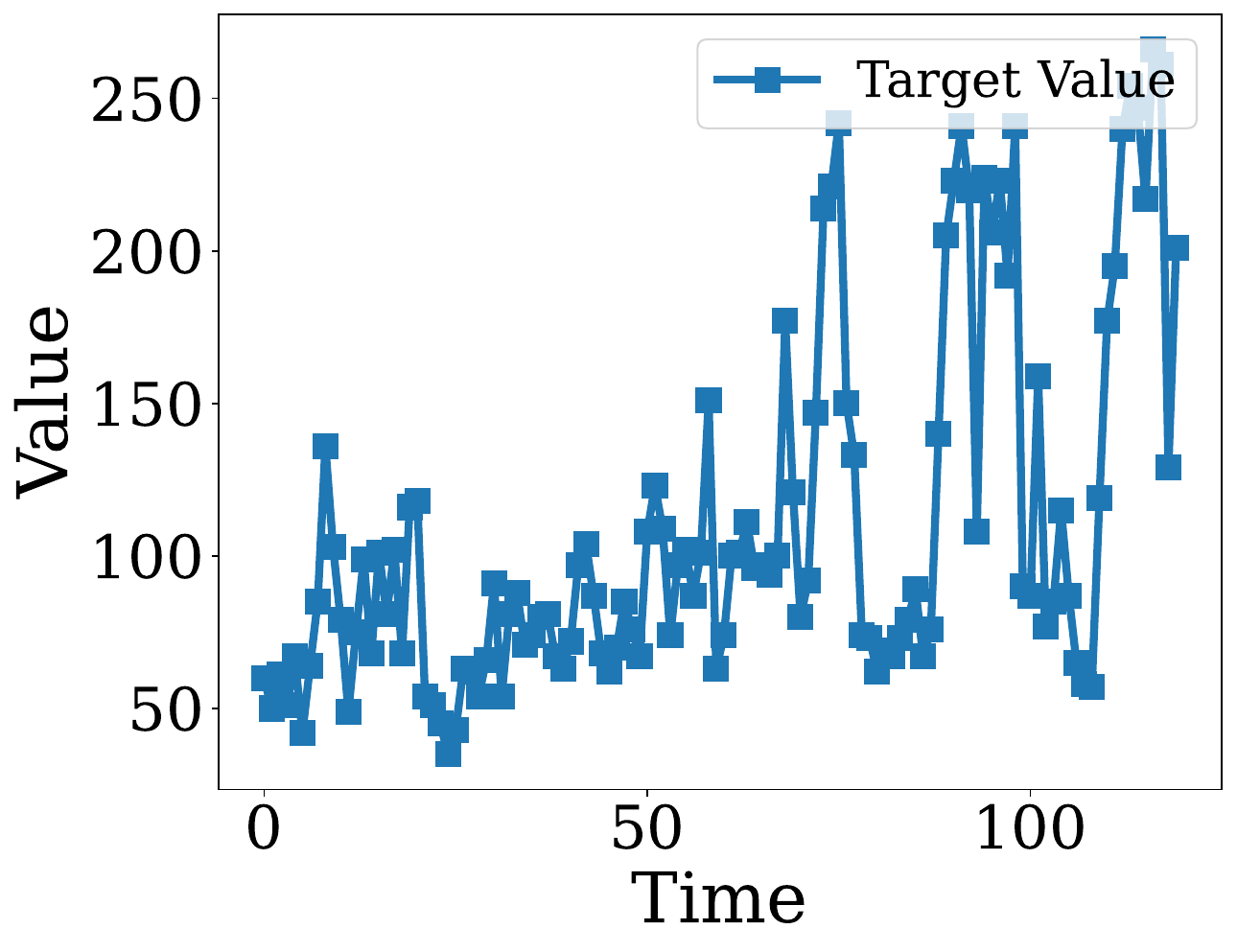}
        \caption{}
    \end{subfigure}
    \caption{More selected time series data from the Environment dataset. The time series data has 120 data points.}
    \label{fig:case_study_ts_environment_data}
\end{figure}

\begin{figure}[t]
    \small
    \centering
    \begin{subfigure}{0.49\linewidth}   
        \includegraphics[width=0.98\linewidth]{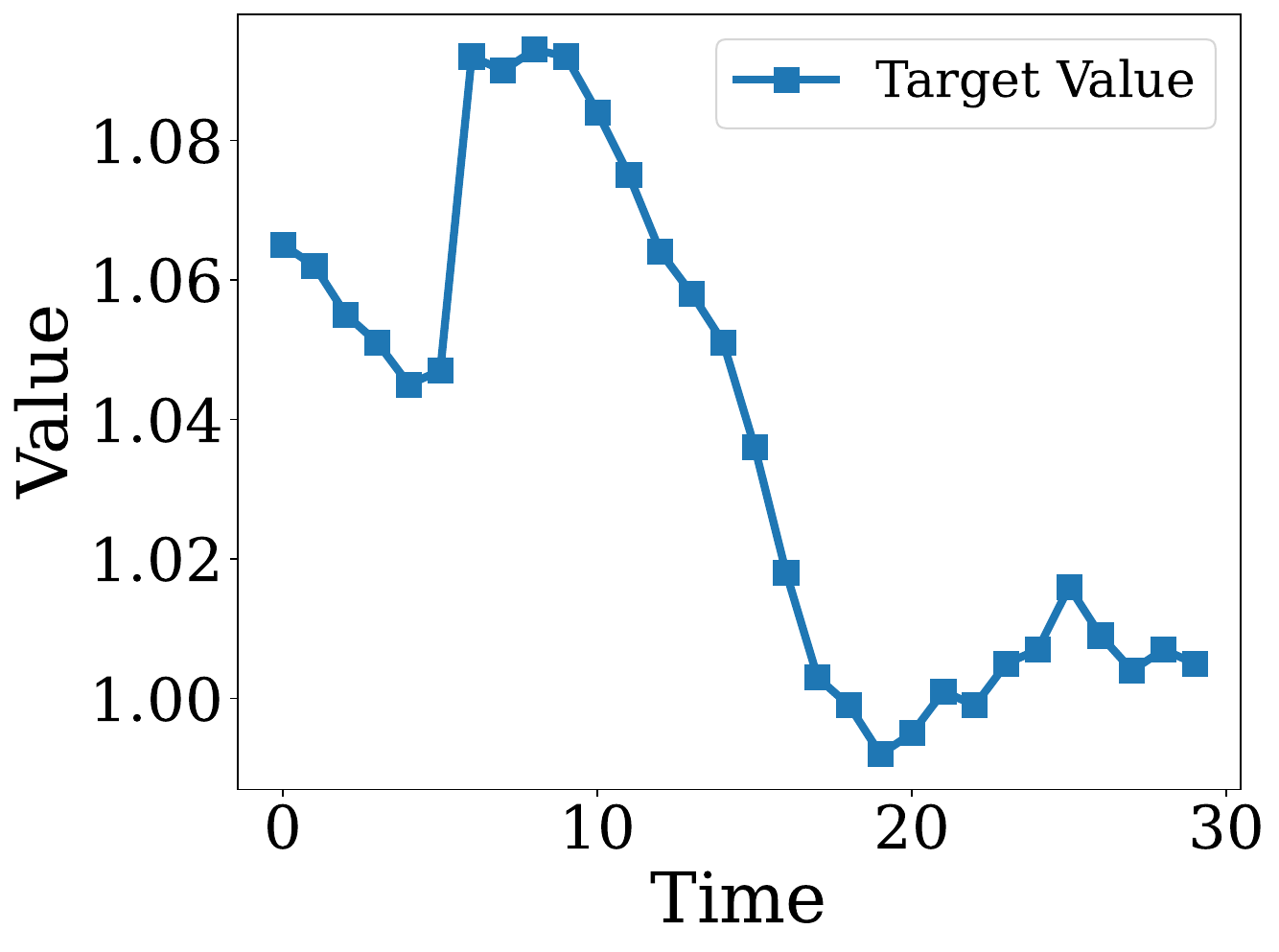}
        \caption{}
    \end{subfigure}
    \begin{subfigure}{0.49\linewidth}   
        \includegraphics[width=0.98\linewidth]{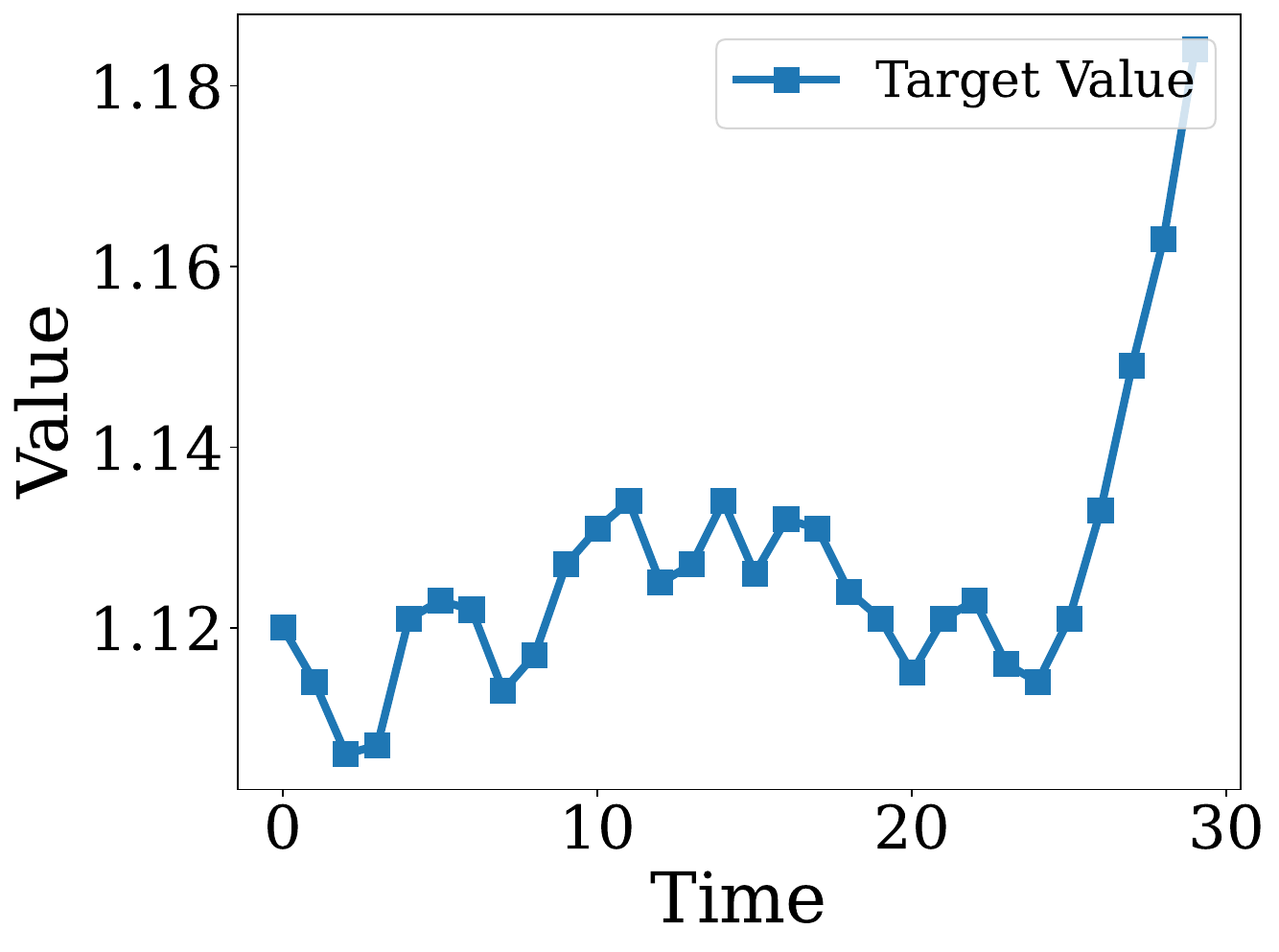}
        \caption{}
    \end{subfigure}
    \caption{More selected time series data from the Energy dataset. The time series data has 36 data points.}
    \label{fig:case_study_ts_energy_data}
\end{figure}

\begin{figure}[t]
    \small
    \centering
    \begin{subfigure}{0.49\linewidth}   
        \includegraphics[width=0.98\linewidth]{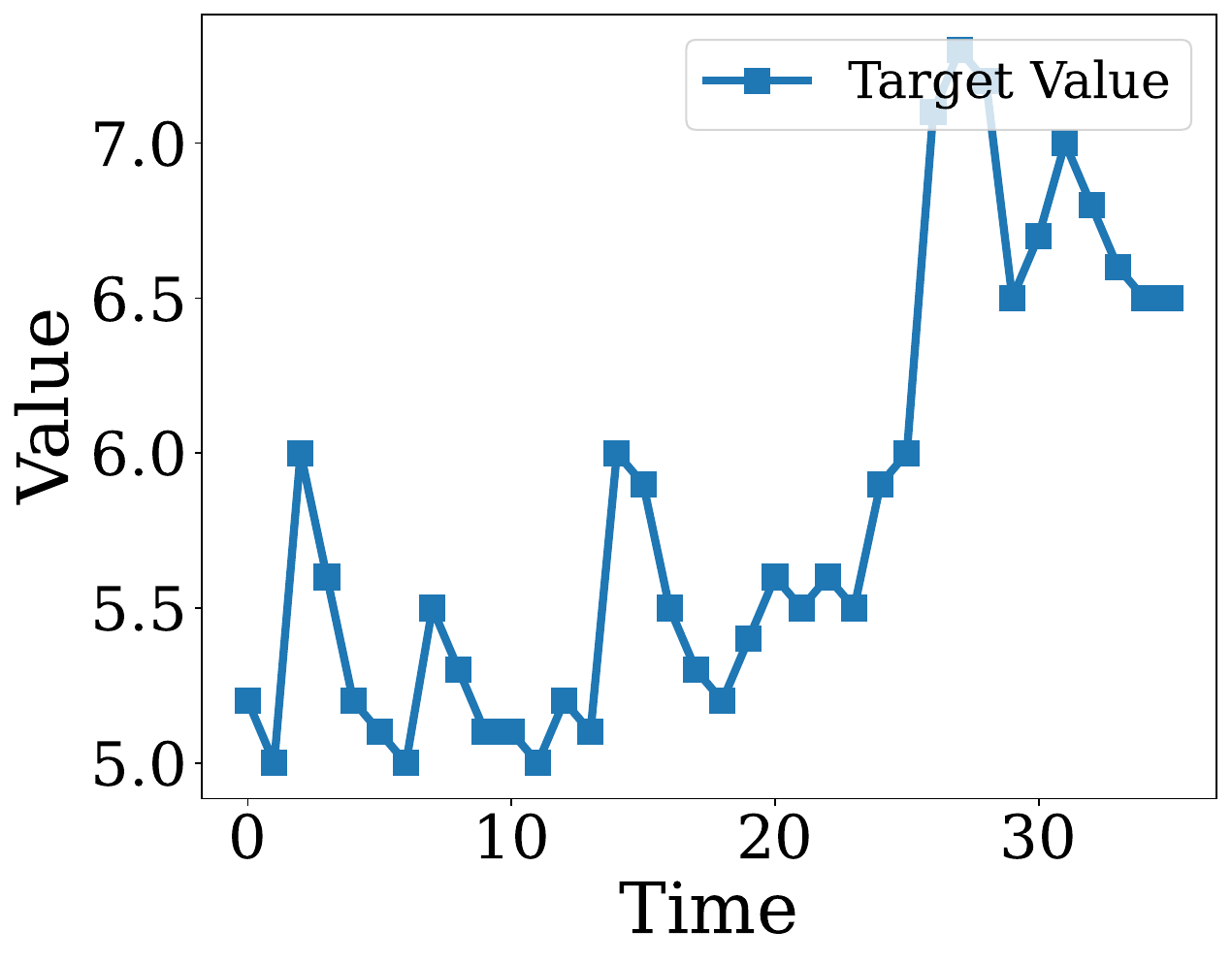}
        \caption{}
    \end{subfigure}
    \begin{subfigure}{0.49\linewidth}   
        \includegraphics[width=0.98\linewidth]{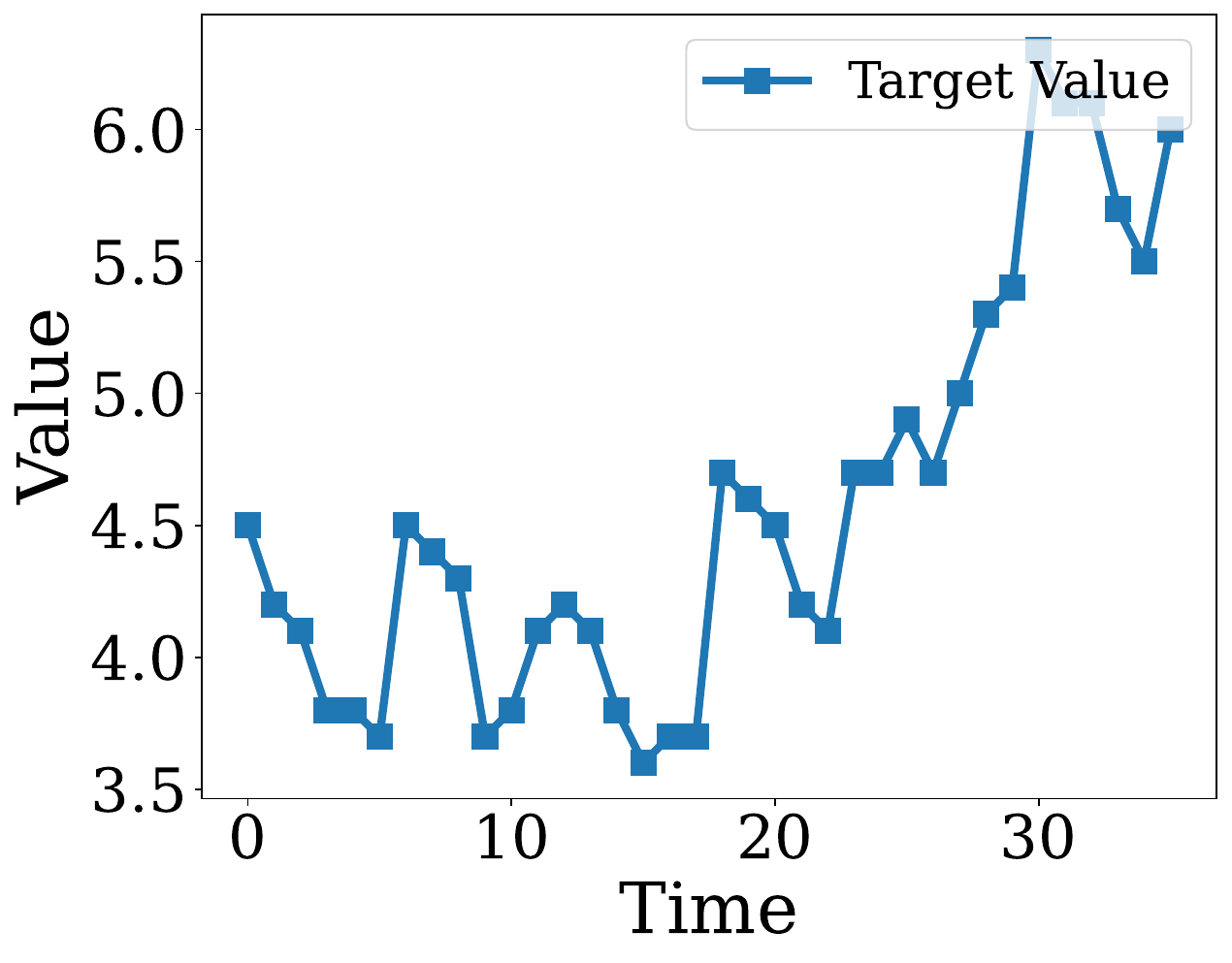}
        \caption{}
    \end{subfigure}
    \caption{More selected time series data from the Social Good dataset. The time series data has 36 data points.}
    \label{fig:case_study_ts_socialgood_data}
\end{figure}

\noindent\textbf{Additional Examples on Various Domains}.
Additional examples are presented for the synthetic, environment, energy and social good datasets, respectively. Specifically, the general annotations of selected time series on the synthetic dataset that in Fig.~\ref{fig:case_study_ts_synthetic_data} are shown in Tables~\ref{tab:case_study_synthetic_more_1} and~\ref{tab:case_study_synthetic_more_2}. The domain-specific annotations of selected time series on the environment dataset (Fig.~\ref{fig:case_study_ts_environment_data}) are shown in Tables~\ref{tab:case_study_environment_more_1} and~\ref{tab:case_study_environment_more_2}. And the domain-specific annotations of time series on the energy dataset (Fig.~\ref{fig:case_study_ts_energy_data}) are shown in Tables~\ref{tab:case_study_energy_more_1} and~\ref{tab:case_study_energy_more_2}. Similarly, the domain-specific annotations of time series on the social good dataset (Fig.~\ref{fig:case_study_ts_socialgood_data}) are shown in Tables~\ref{tab:case_study_socialgood_more_1} and~\ref{tab:case_study_socialgood_more_2}. 
Similar observations to those in Table~\ref{tab:case_study_domain_specific_evaluation_main} and Table~\ref{tab:case_study_synthetic_more_0} are found.


\begin{table*}[t]
\small
\centering
\caption{A case study in Social Good dataset. The target time series data is shown in Fig.~\ref{fig:case_study_ts_data} (a). \colorbox{lightpurple}{Purple} highlights basic patterns, \colorbox{lightgreen}{green} highlights general patterns, and \colorbox{yellow}{yellow} highlights domain-specific patterns.}
\begin{tabular}{p{0.95\linewidth}} \toprule
\textbf{TESSA's general annotations} 

The time series data shows a generally \colorbox{lightpurple}{decreasing trend}, initially rising until a midpoint before consistently declining. \colorbox{lightgreen}{Seasonal} patterns are evident, with regular peaks around specific intervals. The series experiences rapid positive and negative {fluctuations}, reflecting significant changes in the measured system. \colorbox{lightgreen}{Support and resistance levels} are observed, with values frequently pausing near 7.0 and struggling to exceed 10.5. Past values have an impact on current data, indicating that recent patterns help predict future outcomes.


\textbf{DirectLLM's general annotations} 

The uni-variate time series data exhibits \colorbox{lightpurple}{fluctuating} values with an overall \colorbox{lightpurple}{descending trend}. Initially, the data is relatively stable around the 9.5 to 10.5 range, peaking at 11.4 before descending gradually. Midway through the series, after reaching a peak, there is a noticeable decline as values drop from approximately 11 to levels around 8. By the end, the values stabilize somewhat but decline further, reaching a low near 7. This pattern suggests an initial increase followed by a \colorbox{lightpurple}{downward trend} and final stabilization at lower levels, indicating a potential cycle or shift in the underlying phenomenon being measured.

\\ \midrule

\textbf{TESSA's domain-specific annotations} 

The unemployment rate shows a pattern consistent with trends in \colorbox{yellow}{labor market resilience} and economic momentum. Initially stable, with rates between 9 and 10, there is a gradual upward movement indicating early signs of economic growth and potential tightening in the labor market, peaking around 11. This is followed by a downturn, reflecting possible \colorbox{yellow}{macroeconomic constraints} or shifts that reduce \colorbox{yellow}{economic momentum}, as the rate drops to approximately 7.5 towards the end of the period. The rates around 9 and 11 can be seen as \colorbox{lightgreen}{support and resistance} thresholds for employment, where the market experiences inertia. Notably, the subtle \colorbox{lightgreen}{seasonal fluctuations} suggest the presence of moderate cyclic factors influencing employment \colorbox{lightpurple}{trends}, helping to understand broader economic activity patterns.


\textbf{DirectLLM's domain-specific annotations} 

The time series data represents \colorbox{lightpurple}{fluctuations} in the unemployment rate over a specific period. Initially, the unemployment rate starts quite high, around 9.5\%, and exhibits \colorbox{lightpurple}{slight ups and downs} but generally remains above 9\% until a significant increase is observed, reaching its peak at 11.4\%. Following this peak, the rate gradually declines, showing an overall \colorbox{lightpurple}{downward trend} with a notable reduction to as low as 7\%. This decline suggests a potential improvement in the employment situation, perhaps due to economic recovery or effective employment policies. Towards the end of the series, the unemployment rate stabilizes around 7-8\%, although a slight increase is seen again, indicating the dynamic nature of employment conditions.

\\ \bottomrule
\end{tabular}
\label{tab:case_study_main}
\end{table*}

\begin{table*}[t]
\small
\centering
\caption{A case study in Stock dataset. The target \textbf{multivariate} time series data is shown in Fig.~\ref{fig:case_study_ts_multivariate_data} (a). GPT-4o is the LLM backbone. \colorbox{lightpurple}{Purple} highlights basic patterns, \colorbox{lightgreen}{green} highlights general patterns, \colorbox{yellow}{yellow} highlights domain-specific patterns and \colorbox{lightblue}{blue} highlights correlations between variables in multivariate time series data.} 
\begin{tabular}{p{0.95\linewidth}} \toprule 

\textbf{TESSA's domain-specific annotations}

Compass Digital Acquisition Unit's stock price shows a notable pattern of \colorbox{lightgreen}{rising and falling periodically}, indicating \colorbox{lightgreen}{seasonality} with stable \colorbox{yellow}{long-term trend}s interrupted by short-term \colorbox{lightpurple}{fluctuations}. There are key \colorbox{yellow}{resistance levels} around intervals 134, 270, and 403, where prices peak before dipping. The stock volume demonstrates significant spikes at specific points, suggesting irregular activity, particularly around values 7000 and 80500, which may indicate \colorbox{yellow}{volume bursts} or unusual market events. The relative strength index (RSI) also reveals a \colorbox{lightgreen}{recurring} pattern, gradually \colorbox{lightpurple}{trending upward} before a sharp decline, reflecting a cycle of growth and subsequent drop. Overall, the mild \colorbox{lightblue}{positive correlation between stock price and RSI} indicates that periodic changes in price are somewhat echoed in RSI patterns, potentially offering predictive insights for future stock movements.


\textbf{DirectLLM's domain-specific annotations} 

The provided time series data consists of three primary features: price, volume, and relative strength index (RSI). Over the observation period, the price demonstrates an overall \colorbox{lightpurple}{upward trend}, starting around \$9.74, exhibiting \colorbox{lightpurple}{fluctuations}, and rising to hover around \$10.81 towards the end. Notable price spikes correspond with significant \colorbox{lightpurple}{increases} in trading volume, indicating periods of high trading activity, such as jumps to 80,500 and 100,200 in volume. Additionally, the RSI values \colorbox{lightpurple}{range sharply}, highlighting areas of overbought conditions (RSI approaching or at 100) and oversold conditions (RSI dropping around or below 50). These RSI changes suggest periods of potential buying or selling pressure, mirroring the observed price moves.

\\ \bottomrule
\end{tabular}
\label{tab:case_study_multivariate}
\end{table*}

\begin{table*}[t]
\small
\centering
\caption{A case study in the synthetic dataset. The selected time series data is shown in Fig.~\ref{fig:case_study_ts_synthetic_data_main}, where the time series data has 120 data points. \colorbox{lightpurple}{Purple} highlights basic patterns, \colorbox{lightgreen}{green} highlights general patterns, and \colorbox{lightred}{red} highlights hallucinations.}
\begin{tabular}{p{0.95\linewidth}} \toprule
\textbf{TESSA's general annotations} 

This time series exhibits notable fluctuations with values ranging from 35.0 to 118.0. The trend indicates a general \colorbox{lightpurple}{decline}, starting around 65.26 and ending around 53.58 reflecting a significant \colorbox{lightpurple}{downward movement} over time. \colorbox{lightgreen}{Periodic peaks} are captured by the \colorbox{lightgreen}{rolling window feature } showing intermittent spikes up to 118.0, suggesting instances of strong \colorbox{lightgreen}{breakout behavior} despite an overall \colorbox{lightgreen}{support} decrease. The \colorbox{lightgreen}{seasonality} component reveals cyclical patterns with both positive and negative offsets, indicating underlying periodic forces affecting the data, akin to an oscillator. The alignment between the lag feature and the actual values underscores a \colorbox{lightgreen}{historical reliance} or \colorbox{lightgreen}{resilience} of the present data points on past values. reflecting consistency in patterns.


\textbf{DirectLLM's general annotations} 

The provided time series data consists of \colorbox{lightred}{138 values} and exhibits significant fluctuations. There are notable \colorbox{lightpurple}{peaks} at several points such as 118.0 and 113.0, indicating occasional high spikes in values. Conversely, values also \colorbox{lightpurple}{drop} to lows around 35.0, suggesting substantial variability. Most data points seem to oscillate around a mid-range, between approximately 50.0 and 70.0. This pattern suggests periods of \colorbox{lightpurple}{stability} interspersed with sporadic increases and decreases. Analyzing such a dataset can be indicative of a highly dynamic system or process with frequent transitions between states of varying magnitude.

\\ \bottomrule
\end{tabular}
\label{tab:case_study_synthetic_more_0}
\end{table*}

\begin{table*}[t]
\small
\centering
\caption{One more example of general annotation generation in the synthetic dataset. The selected dataset is shown in Fig.~\ref{fig:case_study_ts_synthetic_data} (a). \colorbox{lightpurple}{Purple} highlights basic patterns, \colorbox{lightgreen}{green} highlights general patterns, and \colorbox{lightred}{red} highlights hallucinations. DirectLLM only captures on the basic trend pattern for time series annotation, while TESSA considers more important general patterns.}
\begin{tabular}{p{0.95\linewidth}} \toprule
\textbf{TESSA's general annotations} 
The time series exhibits a complex behavior with an overall \colorbox{lightpurple}{downward trend}, accompanied by periods of fluctuation and temporary upward corrections. Key points include \colorbox{lightgreen}{support and resistance levels}, where the trend either pauses or reverses, notably around values 10.834 and 8.929. Increased \colorbox{lightgreen}{volatility} is evident, indicating active trading periods, which align with higher variability. \colorbox{lightpurple}{Trend} channels are apparent, where the data moves within upper and lower boundaries, particularly showing both descending and emerging upward trends. In some instances, \colorbox{lightgreen}{breakout thresholds} highlight significant changes, signaling \colorbox{lightgreen}{momentum shifts}.


\textbf{DirectLLM's general annotations} 
This uni-variate time series data exhibits a pattern with multiple phases of \colorbox{lightpurple}{rise and fall}, indicative of periodic \colorbox{lightpurple}{fluctuations} over time. Initially, the series starts at a moderate level, gradually ascending to a peak around the values of \colorbox{lightred}{7.58 and 7.52} before experiencing a gradual decline. The data then showcases another rise peaking just above 9 and 10 marks, followed by a sharp and continuous decline, entering negative territory around the value of -6.270 and \colorbox{lightred}{continuing downwards}. Near the end of the series, escalating towards positive values and climaxing at 5.091. The overall structure suggests well-defined periodic or seasonal trends, potentially influenced by external or inherent factors.

\\ \bottomrule
\end{tabular}
\label{tab:case_study_synthetic_more_1}
\end{table*}

\begin{table*}[t]
\small
\centering
\caption{One more example of general annotation generation in the synthetic dataset. The selected dataset is shown in Fig.~\ref{fig:case_study_ts_synthetic_data} (b). \colorbox{lightpurple}{Purple} highlights basic patterns, \colorbox{lightgreen}{green} highlights general patterns, and \colorbox{lightred}{red} highlights hallucinations. DirectLLM only captures on the basic trend pattern for time series annotation, while TESSA considers more important general patterns.}
\begin{tabular}{p{0.95\linewidth}} \toprule
\textbf{TESSA's general annotations} 
The time series data illustrates an overall \colorbox{lightpurple}{upward trend} characterized by increasing values over time, signifying growth. Initially, the series displays stability with minor fluctuations, often not dropping below certain \colorbox{lightgreen}{support levels}, indicating consistency. As the series advances, \colorbox{lightgreen}{breakthrough points} become apparent, where values exceed previous \colorbox{lightgreen}{resistance levels}, suggesting heightened \colorbox{lightgreen}{momentum} and possible shifts in liquidity that drive this progress. The later parts of the data feature more pronounced scending trends, pointing to a stronger \colorbox{lightpurple}{upward movement}. \colorbox{lightgreen}{Periodic} trend reversals also appear, reflecting temporary changes before returning to the \colorbox{lightpurple}{dominant upward trend}, which highlights the series' dynamic nature and potential for \colorbox{lightpurple}{fluctuations}.

\textbf{DirectLLM's general annotations} 
The given uni-variate time series data exhibits a \colorbox{lightpurple}{fluctuating} pattern with a general \colorbox{lightpurple}{upward trend}. Initially, from the first data point, there is a notable \colorbox{lightpurple}{increase} in values, reaching a peak around the seventh value. This is followed by a gradual \colorbox{lightpurple}{decline} and subsequent \colorbox{lightred}{stabilization with minor fluctuations} between the 12th and 31st values. Notably, around the 84th value, a significant surge in values begins, culminating in a prominent \colorbox{lightpurple}{steep increase} towards the end of the series, suggesting a potential exponential growth or shift occurring in the data. Overall, the time series transitions from more stable periods into a pronounced upward trend, signaling potential external influences or underlying factors driving the increase.

\\ \bottomrule
\end{tabular}
\label{tab:case_study_synthetic_more_2}
\end{table*}

\begin{table*}[t]
\small
\centering
\caption{One more example of domain-specific annotation generation in the Environment dataset. The selected dataset is shown in Fig.~\ref{fig:case_study_ts_environment_data} (a). \colorbox{lightpurple}{Purple} highlights basic patterns, \colorbox{lightgreen}{green} highlights general patterns, and \colorbox{yellow}{yellow} highlights the domain-specific patterns.}
\begin{tabular}{p{0.95\linewidth}} \toprule
\textbf{TESSA's domain-specific annotations} 

The air quality index (AQI) data exhibits \colorbox{lightpurple}{significant fluctuations}, with values ranging from 53 to 235 over time, indicating variability in air quality. \colorbox{lightgreen}{Support levels} around values like 100 and 140 suggest periods when air quality temporarily stabilizes or improves. On the other hand, \colorbox{lightgreen}{resistance levels} near values like 200 and 235 show points where air quality struggles to improve further before worsening. Several distinct upward trends, particularly from AQI values 70 to 150 and 177 to 235, indicate temporary periods of improvement in air quality, whereas downward trends around values 166 to 123 and 208 to 84 reflect deteriorating air quality after peaks. Monitoring these trends and critical thresholds will be essential for identifying and responding to significant pollution events effectively.

\textbf{DirectLLM's domain-specific annotations} 

The time series data represents the air quality index (AQI) over a series of observations, showing \colorbox{lightpurple}{fluctuations} in air pollution levels. Initially, the AQI values are \colorbox{lightpurple}{moderate}, transitioning to higher levels, peaking at alarming numbers such as 235 and 209, which indicate very unhealthy air quality. This indicates potential spikes in pollution that could be associated with environmental events or increased urban activity. Periods of lower AQI values suggest moments of improved air quality, but these are often followed by \colorbox{lightpurple}{sharp increases}, highlighting the inconsistency and poor air conditions in the observed timeframe. Overall, the data reflects significant air quality concerns, emphasizing the need for monitoring and potential interventions to safeguard public health.

\\ \bottomrule
\end{tabular}
\label{tab:case_study_environment_more_1}
\end{table*}

\begin{table*}[t]
\small
\centering
\caption{One more example of domain-specific annotation generation in the Environment dataset. The selected dataset is shown in Fig.~\ref{fig:case_study_ts_environment_data} (b). \colorbox{lightpurple}{Purple} highlights basic patterns, \colorbox{lightgreen}{green} highlights general patterns, and \colorbox{yellow}{yellow} highlights the domain-specific patterns.}
\begin{tabular}{p{0.95\linewidth}} \toprule
\textbf{TESSA's domain-specific annotations} 

The air quality index (AQI) exhibits \colorbox{lightpurple}{noticeable upward and downward trends} over time, indicating periods of improvement and decline in air quality. Regular \colorbox{lightgreen}{seasonal patterns} are apparent, with AQI values \colorbox{lightgreen}{cyclically} rising and falling, suggesting that certain times of the year may be more prone to higher pollution levels. \colorbox{lightgreen}{Critical thresholds} in AQI levels highlight points where air quality significantly improves or deteriorates, with some values acting as \colorbox{lightgreen}{resistance levels} that AQI struggles to surpass. Episodes of \colorbox{yellow}{high variability} in AQI indicate times of significant fluctuations, potentially due to varying pollution sources or climate conditions. Lastly, \colorbox{lightgreen}{breakout patterns} where AQI values suddenly shift indicate potential changes in pollution control effectiveness or new influences on air quality dynamics.


\textbf{DirectLLM's domain-specific annotations} 

The provided time series data represents \colorbox{lightpurple}{fluctuations} in the air quality index (AQI), with values indicating varying levels of air pollution over time. Initially, the AQI remains in a moderate range, typically below 100, but there is a noticeable \colorbox{lightpurple}{spike} as values reach up to 220, indicating very poor air quality. This suggests that certain periods experienced significantly higher pollution levels, which can have serious implications for public health and environmental quality. The data shows some improvements, but the \colorbox{lightpurple}{peaks raise} concerns about excessive pollution, particularly during certain months. Continuous monitoring and mitigation of pollution sources are essential to reduce the frequency and severity of these dangerous air quality levels.


\\ \bottomrule
\end{tabular}
\label{tab:case_study_environment_more_2}
\end{table*}

\begin{table*}[t]
\small
\centering
\caption{One more example of domain-specific annotation generation in the Energy dataset. The selected dataset is shown in Fig.~\ref{fig:case_study_ts_energy_data} (a). \colorbox{lightpurple}{Purple} highlights basic patterns, \colorbox{lightgreen}{green} highlights general patterns, and \colorbox{yellow}{yellow} highlights the domain-specific patterns.}
\begin{tabular}{p{0.95\linewidth}} \toprule
\textbf{TESSA's domain-specific annotations} 

Gas prices have exhibited an upward trend, peaking at \$1.077 before experiencing a slight decline towards the latter part of the series. This indicates an initial growth phase followed by stabilization. Minor, yet consistent \colorbox{yellow}{seasonal variations} suggest \colorbox{yellow}{periodic influences} that slightly affect the price. The analysis of residuals reveals small but steady fluctuations, indicating some level of unpredictability in the market. \colorbox{lightgreen}{Support and resistance} analysis identifies \colorbox{lightgreen}{critical thresholds} around \$1.003 and \$1.092, highlighting potential \colorbox{yellow}{reversal points} and \colorbox{yellow}{consolidation zones}.

\textbf{DirectLLM's domain-specific annotations} 

The time series data represents \colorbox{lightpurple}{fluctuations} in gas prices over a period of time, listed in \colorbox{lightpurple}{ascending order.} Initially, the gas prices start around \$1.06 but gradually show a \colorbox{lightpurple}{downward trend}, reaching as low as approximately \$0.99 before slightly rising again. This decline suggests that there may have been favorable market conditions or increased supply during this period, which typically leads to lower prices for consumers. Toward the end of the observed period, prices stabilize around the \$1.00 mark, indicating a potential plateau or gradual recovery in price levels. Overall, the data illustrates a dynamic market scenario, highlighting the volatility and changing nature of gas prices influenced by various factors in the energy domain.

\\ \bottomrule
\end{tabular}
\label{tab:case_study_energy_more_1}
\end{table*}

\begin{table*}[t]
\small
\centering
\caption{One more example of domain-specific annotation generation in the Energy dataset. The selected dataset is shown in Fig.~\ref{fig:case_study_ts_energy_data} (b). \colorbox{lightpurple}{Purple} highlights basic patterns, \colorbox{lightgreen}{green} highlights general patterns, and \colorbox{yellow}{yellow} highlights the domain-specific patterns.}
\begin{tabular}{p{0.95\linewidth}} \toprule
\textbf{TESSA's domain-specific annotations} 

Gas prices show a generally \colorbox{lightpurple}{upward trend}, starting around 1.11 and gradually increasing over time, indicating a \colorbox{yellow}{long-term positive movement}. Minor \colorbox{lightgreen}{seasonal fluctuations} are observed, but they do not dominate the overall trend. There are \colorbox{lightgreen}{key support levels} around 1.113 and \colorbox{lightgreen}{resistance levels} at 1.133, which may act as \colorbox{yellow}{pivotal points} for future price movements. Residuals suggest occasional minor deviations from the trend due to random factors, resulting in a generally stable series with \colorbox{yellow}{slight intraday volatility}. The \colorbox{lightgreen}{moving average} values reinforce this steady climb, suggesting continued \colorbox{lightpurple}{stability} with periodic minor disruptions in gas prices.

\textbf{DirectLLM's domain-specific annotations} 

The time series data represents \colorbox{lightpurple}{fluctuations} in gas prices over a specific period, with values consistently hovering around the \$1.11 to \$1.18 range. Notably, the data shows slight \colorbox{lightpurple}{increases and decreases} in price, suggesting moderate volatility within this timeframe. The highest recorded price peaks at approximately \$1.18, reflecting potential market adjustments or external factors influencing gas prices. This stability could indicate a balanced supply and demand scenario, although one should remain aware that various events, such as geopolitical developments, can lead to sudden shifts. Overall, these insights provide a snapshot of gas price trends, useful for consumers and industry stakeholders in making informed decisions related to energy expenditures.
\\ \bottomrule
\end{tabular}
\label{tab:case_study_energy_more_2}
\end{table*}

\begin{table*}[t]
\small
\centering
\caption{One more example of domain-specific annotation generation in the Social Good dataset. The selected dataset is shown in Fig.~\ref{fig:case_study_ts_socialgood_data} (a). \colorbox{lightpurple}{Purple} highlights basic patterns, \colorbox{lightgreen}{green} highlights general patterns, and \colorbox{yellow}{yellow} highlights the domain-specific patterns.}
\begin{tabular}{p{0.95\linewidth}} \toprule
\textbf{TESSA's domain-specific annotations} 

The unemployment rate data shows a gentle \colorbox{lightpurple}{upward trend} from approximately 5.4\% to 6.57\% over the observed period, indicating a \colorbox{lightpurple}{consistent rise} in unemployment levels. This gradual increase implies a \colorbox{yellow}{weakening labor market}, which could be influenced by broader socioeconomic conditions. Although there are oscillations suggesting \colorbox{lightgreen}{cyclical} patterns, these are not strong and exhibit some \colorbox{yellow}{irregularities}, pointing to potential \colorbox{yellow}{short-term} fluctuations or \colorbox{yellow}{external disruptions}. The stable relationship between consecutive data points suggests that the unemployment rate changes are relatively steady without abrupt shifts, allowing for some degree of predictability. Overall, the trend may signal policymakers to consider implementing measures to address the rising unemployment, especially if the current trajectory continues.

\textbf{DirectLLM's domain-specific annotations} 
The time series data represents the unemployment rate over a period of time. Initially, the unemployment rate showed a \colorbox{lightpurple}{slight fluctuation} around the 5\% mark, indicating a relatively stable job market. However, a noticeable increase began, with rates rising to over 7\%, suggesting a worsening economic situation and potential job losses. After peaking at 7.3\%, the data indicates a \colorbox{lightpurple}{gradual decrease}, though the rates remained relatively high around 6.5\% towards the end. This pattern highlights a period of economic challenge, with gradual steps towards recovery.

\\ \bottomrule
\end{tabular}
\label{tab:case_study_socialgood_more_1}
\end{table*}

\begin{table*}[t]
\small
\centering
\caption{One more example of domain-specific annotation generation in the Social Good dataset. The selected dataset is shown in Fig.~\ref{fig:case_study_ts_socialgood_data} (b). \colorbox{lightpurple}{Purple} highlights basic patterns, \colorbox{lightgreen}{green} highlights general patterns, and \colorbox{yellow}{yellow} highlights the domain-specific patterns.}
\begin{tabular}{p{0.95\linewidth}} \toprule
\textbf{TESSA's domain-specific annotations} 
The unemployment rate data reflects a positive \colorbox{yellow}{economic momentum}, as evidenced by a \colorbox{lightpurple}{persistent upward trend} starting from around 4.1 and moving above 6, suggesting a gradual increase in unemployment over the series. This trend could indicate rising employment participation pressure or an effectiveness of \colorbox{yellow}{policy breakthrough effects} aimed at fostering job growth, which could be interpreted as a \colorbox{yellow}{counter-effect} if not analyzed in tandem with job creation data. Despite the \colorbox{lightpurple}{upward trend}, the series shows \colorbox{lightgreen}{seasonality} patterns, with \colorbox{lightgreen}{periodic} fluctuations above and below a baseline, suggesting regular influences on employment rates, possibly due to economic cycles or policy adjustments. These fluctuations may align with safety net thresholds, which temporarily reduce unemployment rates during specific intervals. In summary, the data represents a complex interplay of economic dynamics, signaling opportunities for policy interventions to maintain a balance between growth and employment stability.

\textbf{DirectLLM's domain-specific annotations} 

The time series data illustrates the \colorbox{lightpurple}{fluctuation} of the unemployment rate over a specific period. Initially, the unemployment rate started at a level of 4.5\% and experienced a \colorbox{lightpurple}{gradual decline} to a low of about 3.6\%. However, this was followed by an \colorbox{lightpurple}{upward trend}, eventually reaching highs of around 6.3\%. Such \colorbox{lightpurple}{fluctuations} might have been influenced by changing economic conditions, labor market policies, or external events impacting employment. Notably, the trend indicates periods of economic strengthening followed by downturns, reflecting possible cycles of growth and contraction in the job market.

\\ \bottomrule
\end{tabular}
\label{tab:case_study_socialgood_more_2}
\end{table*}






